\documentclass[10pt,twocolumn,letterpaper]{article}
\pdfoutput=1
\usepackage{wacv}
\usepackage{times}
\usepackage{epsfig}
\usepackage{graphicx}
\usepackage{amsmath}
\usepackage{amssymb}
\usepackage{subcaption}
\usepackage{comment}
\makeatletter
\@namedef{ver@everyshi.sty}{}
\makeatother
\usepackage{enumitem}
\usepackage{pgfplots}
\usepackage{multirow}
\usepackage{booktabs}
\usepackage{xcolor,colortbl}

\usepackage{algorithm}
\usepackage{algorithmic}
\usepackage{calc}
\usepackage{needspace}
\newcolumntype{+}{>{\global\let\currentrowstyle\relax}}
\newcolumntype{-}{>{\currentrowstyle}}

\wacvfinalcopy 

\def\wacvPaperID{109} 

\ifwacvfinal\fi
\begin{document}

\title{Boosting Standard Classification Architectures Through a Ranking Regularizer}


\author{Ahmed Taha$^1$
\quad
Yi-Ting Chen$^2$
\quad
Teruhisa Misu$^2$
\quad 
Abhinav Shrivastava$^1$
\quad 
Larry Davis$^1$
\\
$^1$University of Maryland, College Park \qquad $^2$Honda Research Institute, USA\\
}



\maketitle
\ifwacvfinal\thispagestyle{empty}\fi

\begin{abstract}
We employ triplet loss as a feature embedding regularizer to boost classification performance. Standard architectures, like ResNet and Inception, are extended to support both losses with minimal hyper-parameter tuning. This promotes generality while fine-tuning pretrained networks. Triplet loss is a powerful surrogate for recently proposed embedding regularizers. Yet, it is avoided due to large batch-size requirement and high computational cost. Through our experiments, we re-assess these assumptions.

During inference, our network supports both classification and embedding tasks without any computational overhead. Quantitative evaluation highlights a steady improvement on five fine-grained recognition datasets.  Further evaluation on an imbalanced video dataset achieves significant improvement. Triplet loss brings feature embedding capabilities like nearest neighbor to classification models. Code available  at
 \textit{http://bit.ly/2LNYEqL}

%

\end{abstract}

\section{Introduction}
\begin{figure}
		\centering
	\begin{subfigure}[b]{0.24\textwidth}
		\includegraphics[width=\textwidth]{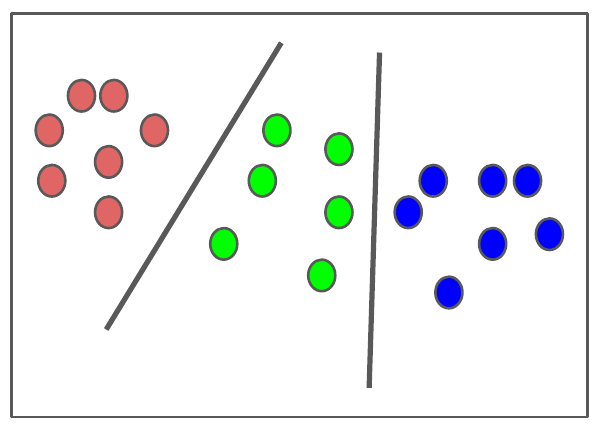}
		\caption{Softmax Loss}
		\label{fig:intro_softmax}
	\end{subfigure}\begin{subfigure}[b]{0.24\textwidth}
	\includegraphics[width=\textwidth]{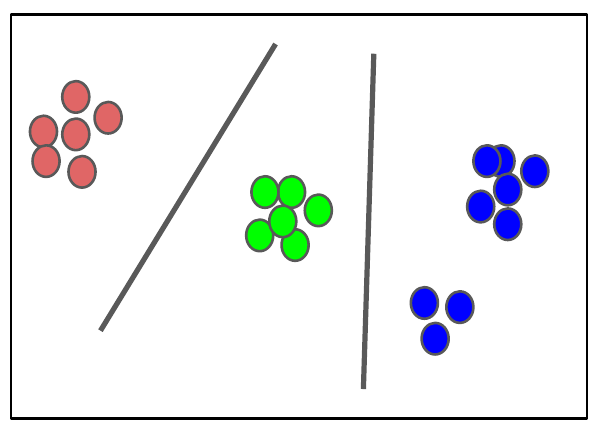}
	\caption{Triplet Loss Regularizer}
	\label{fig:intro_triplet_loss}
\end{subfigure}
\caption{Softmax learns powerful representations with limited embedding regularization. Triplet loss promotes better embedding without an explicit number of class centers.}
\label{fig:intro_section}
\end{figure}

Standard convolutional architectures~\cite{he2016deep,szegedy2015going} learn powerful representation for classification. Pretrained ImageNet~\cite{deng2009imagenet} weights scale their strength through fine tuning to novel domains and relax the large labeled dataset requirement. Yet, the learned representation through softmax attains limited intra-class compactness and inter-class separation. To advocate for a better embedding quality, we propose a two-head architecture. We leverage triplet loss~\cite{schroff2015facenet} as a classification regularizer. It promotes a better feature embedding by attracting similar and repelling different classes as shown in Figure~\ref{fig:intro_section}. This embedding also raises classification model interpretability by enabling nearest neighbor retrieval.

Embedding losses have been successfully applied in conjunction with  softmax loss as regularizers. For example, center loss~\cite{wen2016discriminative} was proposed for better face recognition efficiency. Magnet loss~\cite{rippel2015metric} generalizes the unimodality assumption of center loss. A recent triplet-center loss (TCL)~\cite{he2018triplet} uses only a unimodal embedding but introduced a repelling force between class centers, \ie, inter-class margin maximization. All these methods assume a fixed number of class centers (embedding modes) for all classes.

\begin{figure*}[t!]
	\begin{center}
		\includegraphics[width=1.0\linewidth]{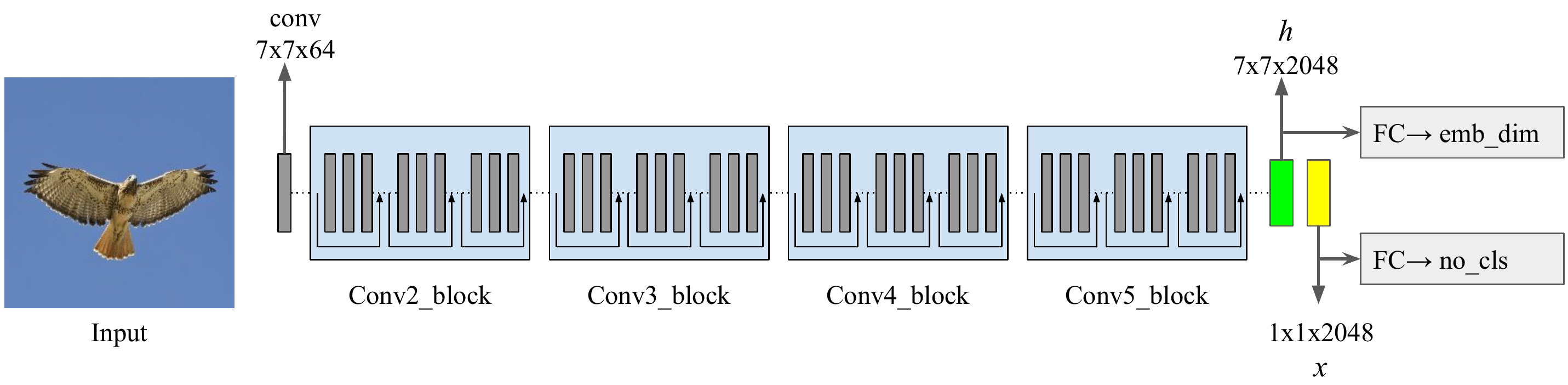}
	\end{center}
	\caption{Our proposed two-head architecture builds on standard networks -- ResNet used for visualization, $x_{input} = pool(h_{input})$. Besides computing classification logits,  the pre-logits layer supports the embedding head. Softmax and triplet losses are applied to the classification logits and embedding features, respectively.}
		\label{fig:arch}
\end{figure*}
Unlike the aforementioned approaches, the standard triplet loss requires no explicit number of embedding modes. Thus, it avoids computing class centers while promoting intra-class compactness and inter-class margin maximization. Surprisingly, recent papers~\cite{wen2016discriminative,he2018triplet} do \textit{not} report the softmax+triplet loss quantitative evaluation. Assumptions about large training batch requirement~\cite{schroff2015facenet} for faster convergence or high batch-processing complexity, to compute pairwise distance matrix, have hindered triplet loss's adoption. Our experiments reassess these assumptions through multiple triplet loss sampling strategies.

To incorporate embedding losses, previous approaches employ loss-specific architectures. This custom setting is imperfect for the softmax baseline as it omits the pre-trained ImageNet weights.  
Through our proposed seamless integration into standard CNNs, we push our baselines' limits. We introduce an embedding head similar to the classification head. Each head applies a single fully connected (FC) layer on the pre-logit convolutional layer features. Figure~\ref{fig:arch} shows our two head architecture where the  pre-logit convolutional features support both softmax and triplet losses for classification and embedding respectively.  This integration boosts classification performance while promoting better embedding.



We evaluate our approach on various classification domains. The first  is a fine-grained visual recognition (FGVR) across five datasets. The second domain is an ego-motion action recognition task with high class imbalance. Large improvements (1-4\%) are achieved in both domains. Evaluation on multiple architectures with the same hyper-parameters highlights our approach's generality. The large batch size requirement represents a key challenge for triplet loss adoption; Schroff \etal~\cite{schroff2015facenet} use a batch-size $b=1800$ and trained on a CPU cluster for 1,000 to 2,000 hours.  In our experiments, we show that using a small batch size $b=32$ still improves performance. A further qualitative evaluation highlights beneficial qualities like nearest neighbor retrieval added to standard classification architectures. 
In summary, the key contributions of this paper are:

\begin{enumerate}[noitemsep]
\item A two-head architecture proposal that uses triplet loss as a regularizer to boost standard architectures' performance through promoting a better feature embedding.
\item A re-evaluation of the large batch size requirement and high computational cost assumptions for triplet loss,
\item Enable better nearest neighbor retrieval on standard classification architectures.
\end{enumerate}

\section{Related Work}

\begin{figure*}
	\centering
	\begin{subfigure}[b]{0.24\textwidth}
		\includegraphics[width=\textwidth]{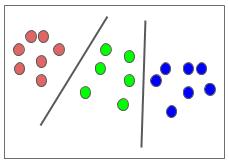}
		\caption{Softmax Loss}
		\label{fig:softmax}
	\end{subfigure}
	\begin{subfigure}[b]{0.24\textwidth}
		\includegraphics[width=\textwidth]{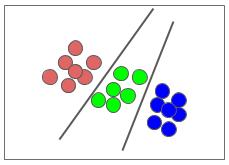}
		\caption{Center Loss Regularizer}
		\label{fig:center}
	\end{subfigure}
	\begin{subfigure}[b]{0.24\textwidth}
		\includegraphics[width=\textwidth]{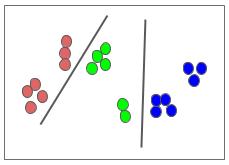}
		\caption{Magnet Loss Regularizer}
		\label{fig:magnet}
	\end{subfigure}
	\begin{subfigure}[b]{0.24\textwidth}
	\includegraphics[width=\textwidth]{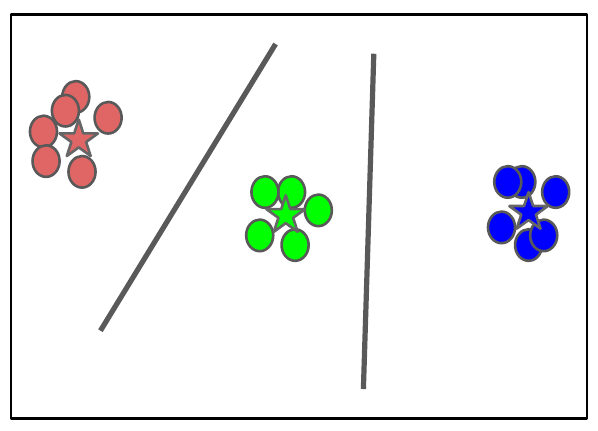}
	\caption{Triplet Center Regularizer}
	\label{fig:tcl}
\end{subfigure}
	\caption{Visualization of softmax and feature embedding regularizers. Softmax separates samples with neither class compactness nor margin maximization considerations. Center loss promotes unimodal compact class while magnet loss supports multi-modal embedding. Triplet center loss strives for unimodal, margin maximization and class compactness. The computed classes' centers are depicted using a star symbol }
	\label{fig:related_work}
\end{figure*}

Visual recognition deep networks employ softmax loss  as follows
\begin{equation}\label{eq:softmax}
L_{ \text{soft} }=-\sum _{ i=1 }^{ b }{ \log { \frac { { e }^{ { W }_{ { y }_{ i } }^{ T }{ x }_{ i } } }{ \sum _{ j=1 }^{ n }{ { e }^{ { W }_{ { j } }^{ T }{ x }_{ i } } }  }  }  }, 
\end{equation}
where  $x_i \in R^{d}$ denotes the $i$th deep feature, belonging to the $y_i$th class. In standard architectures, $x_i$ is the pre-logit layer; the result of flattening the pooled convolutional features as shown in Figure~\ref{fig:arch}. $W_j \in R^d $ denotes the $j$th column of the weights $W \in R^{d\times n}$ in the last fully connected layer.  $b$ and $n$ are the batch size and the number of class respectively. The softmax loss only cares about separating samples from different class. It disregards properties like intra-class compactness and inter-class margin maximization. Embedding regularization is one way to tackle this limitation. Figure~\ref{fig:related_work} depicts different embedding regularizers; all require an explicit number of embedding modes.


\subsection{Center Loss}
Wen \etal~\cite{wen2016discriminative} propose center loss to minimize intra-class variations. By maintaining a per class representative feature vector $c_{yi} \in R^d$, the novel loss term in equation~\ref{eq:center_loss} is proposed. The class centers are computed by averaging corresponding class features. They are updated after every training mini-batch. To avoid perturbations caused by noisy samples, a hyper-parameter $\alpha$ controls the learning rate of the centers, \ie, moving average.
\begin{equation}\label{eq:center_loss}
L_{ \text{cen} }=\frac { 1 }{ 2 } \sum _{ i=1 }^{ b }{ { \parallel x_{ i }-c_{ yi }\parallel  }_{ 2 }^{ 2 } }.
\end{equation}

\subsection{Magnet Loss}
Rippel \etal~\cite{rippel2015metric} propose a center loss term supporting multi-modal embedding, dubbed magnet loss. It computes $K$ class representatives, \ie, K-clusters per class. Each sample is iteratively assigned to one of the K clusters and pushed towards its center. The magnet loss adaptively sculpts the representation space by identifying and enforcing intra-class variation and inter-class similarity. This is formulated as follows

\begin{equation}
\scalebox{0.95}[1]{$L_{ M }=\frac { 1 }{ N } \sum _{ i=1 }^{ N }{ -\log { \frac { \exp ( { \begin{matrix} \frac { -1 }{ 2{ \sigma  }^{ 2 } }  { \parallel { x }_{ i }^{ k }-{ \mu }_{ k }^{ c }\parallel  }_{ 2 }^{ 2 }-\alpha  \end{matrix} }) }{ \sum _{ c\neq C({ x }_{ i }^{ k }) }^{  }{ \sum _{ k=1 }^{ K }{ \exp( { \begin{matrix} \frac { -1 }{ 2{ \sigma  }^{ 2 } }   { \parallel { x }_{ i }^{ k }-{ \mu }_{ k }^{ c }\parallel  }_{ 2 }^{ 2 }-\alpha  \end{matrix} }) }  }  }  }  }$},
\end{equation}
where $N$ and $K$ are the number of samples and clusters per class respectively. $x_i^k \in R^{d}$ denotes the $i$th deep feature, belonging to cluster $k$ in the $y_i$th class, $\mu_{k}^c \in R^{d}$ is the $k$th cluster center belonging to class $c$. Finally ${ \sigma  }^{ 2 }=\frac { 1 }{ N-1 } \sum _{  }^{  }{ { \parallel { x }_{ i }^{ k }-{ \mu  }_{ k }^{ c }\parallel  }_{ 2 }^{ 2 } } $ is the variance of all samples from their respective centers. One criticism of magnet loss is the complexity overhead to maintain multiple clusters per class and their assigned samples. Moreover, the constant number of clusters per-class disputes with imbalanced data distributions.



\subsection{Triplet Center Loss}
While promoting class compactness, the center loss depends on the softmax loss supervision signal to push different classes apart. The learned features optimized with the softmax loss supervision signal are not discriminative enough, \ie, no explicit repelling force pushes different classes apart. Inter-class clusters can overlap due to missing an explicit inter-class repelling incentive. He \etal~\cite{he2018triplet} propose triplet center loss (TCL) to avoid this limitation. By maintaining a per class center $c_{yi} \in R^d$ similar to~\cite{wen2016discriminative}, TCL is formulated as follows
\begin{equation}\label{eq:triplet_center_loss}
L_{ \text{tcl} }=\sum _{ i=1 }^{ b }{ { \left[ { (D({ x }_{ i },{ c }_{ yi })-\min _{ j\neq i }{ D({ x }_{ i },{ c }_{ yj }) } +m) }  \right]  }_{ + }   }, 
\end{equation}
where $m$ is a separating margin, ${ \left[ . \right]  }_{ + } = max(0,.)$  and $D(.)$ represents the squared Euclidean distance function.

Triplet loss is a well-established surrogate for TCL. It achieves the intra and inter-class embedding objectives without computing class centers. Yet, it is largely avoided for its computational complexity and large training batch requirement assumptions. In the experiment section, we address these concerns and evaluate the utility of triplet loss as a regularizer. Our approach is evaluated on the challenging FGVR task where intra-class overwhelm inter-class variations. Further evaluation on the Honda driving dataset (HDD) demonstrates our approach's competence on an imbalanced video dataset. Triplet loss regularization not only lead to higher classification accuracy but also enables better feature embedding.

\section{The Triplet Loss  Regularizer}
The next subsection introduces triplet loss~\cite{schroff2015facenet} as a softmax loss regularizer. Then, we explain our standard architectural extension to integrate an embedding loss.

\subsection{Triplet Loss}
Triplet loss~\cite{schroff2015facenet} has been successfully applied in face recognition~\cite{schroff2015facenet,sankaranarayanan2016triplet} and person re-identification~\cite{cheng2016person,su2016deep,ristani2018features}. In both domains, it is used as a feature embedding tool to measure similarity between objects and provide a metric for clustering. In this work, we utilize triplet loss as a classification regularizer. It is more efficient than contrastive loss~\cite{hadsell2006dimensionality,li2017improving}, and less computationally expensive than  quadruplet~\cite{huang2016local,chen2017beyond} and quintuplet~\cite{huang2016learning} losses. While the pre-logits layer learns better representations for classification using the softmax loss, triplet loss promotes a better feature embedding. Equation~\ref{eq:triplet} shows the triplet loss formulation

\begin{equation}\label{eq:triplet}
L_{ \text{tri} }=\frac{1}{b} \sum _{ i=1 }^{ b }{ { \left[  { (D(a_i,p_i)-{ D(a_i,n_i) } +m) }  \right]  }_{ + }  }, 
\end{equation}
where an anchor image's embedding $a$ of a specific class is pushed closer to a positive image's embedding $p$ from  the same class than it is to a negative image's embedding $n$ of a different class.  Equation~\ref{eq:final} is our loss function with a balancing hyper-parameter $\lambda$.

\begin{equation}\label{eq:final}
L = L_{ \text{soft} } + \lambda L_{ \text{tri} }.
\end{equation}

\noindent\underline{\textbf{Sampling:}} Triplet loss performance is dependent on its sampling strategy. We evaluate both the hard~\cite{hermans2017defense} and semi-hard~\cite{schroff2015facenet} sampling strategies. In semi-hard negative sampling, instead of picking the hardest positive-negative samples, all anchor-positive pairs and their corresponding semi-hard negatives are considered. Semi-hard negatives  satisfy equation~\ref{eq:semi_neg}. They are further away from the anchor than the positive exemplar, yet within the banned margin $m$.  
\begin{equation}\label{eq:semi_neg}
D(a,p) < { D( a,n) } < D( a , p ) + m.
\end{equation}

\begin{figure}[t]
	\centering
	\includegraphics[width=0.5\linewidth]{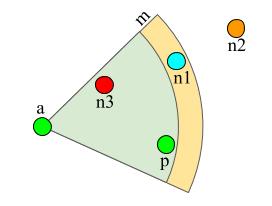}
	\caption{Triplet loss tuple (anchor, positive, negative) and margin $m$. Hard, semi-hard and easy negatives highlighted in red, cyan and orange, respectively.}
	\label{fig:semi_neg}
\end{figure}

\begin{figure}[t]
	\centering
	\includegraphics[width=0.65\linewidth]{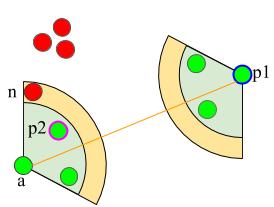}
	\caption{Hard sampling promotes unimodal embedding by picking the farthest positive and nearest negative $(a,p1,n)$. Semi-hard sampling picks $(a,p2,n)$ and avoids any tuple $(a,p,n)$ where $n$ lies between $a$ and $p$.}
	\label{fig:hard_mining}
\end{figure}

Figure~\ref{fig:semi_neg} shows a triplet loss tuple and highlights the different types of negative exemplars: easy ($n_2$), semi-hard ($n_1$) and hard ($n_3$) negatives.
An easy negative satisfies the margin constraint and suffers a zero loss. Unlike hard-sampling, semi-hard sampling supports a multi-modal embedding. Hard sampling picks the farthest positive and nearest negative without any consideration for the margin. In contrast, Figures~\ref{fig:hard_mining} illustrates how semi-hard sampling ignores hard negatives. Two classes, red and green, are embedded into one and two clusters respectively. A hard sampling strategy pulls the farthest positive from one cluster to the anchor in the other cluster, i.e. promotes a merge. The semi-hard sampling strategy omits this tuple because the negative sample is nearer than the positive.

The existence of a semi-hard negative is not guaranteed in small batches, especially near convergence. Thus, we prioritize negative exemplars as illustrated in Figure~\ref{fig:semi_neg}. First priority is given to semi-hard ($n_1$), then easy ($n_2$) and finally hard negatives ($n_3$).


\subsection{Two-Head Architecture}
Standard convolutional architectures, with ImageNet~\cite{deng2009imagenet} weights, are employed in various applications for their powerful representation. 
 We seek to leverage pre-trained standard networks for their advantages in tasks like  fine-grained visual recognition~\cite{lin2015bilinear,krause2016unreasonable,krause2015fine}. This key integration promotes the generality of our approach and distances our work from ~\cite{wen2016discriminative,he2018triplet,sun2015deeply} which use custom architectures. Through experiments, we demonstrate how triplet loss achieves superior classification efficiency compared to center loss.


\begin{figure}[t]
	\centering
	\includegraphics[width=0.7\linewidth]{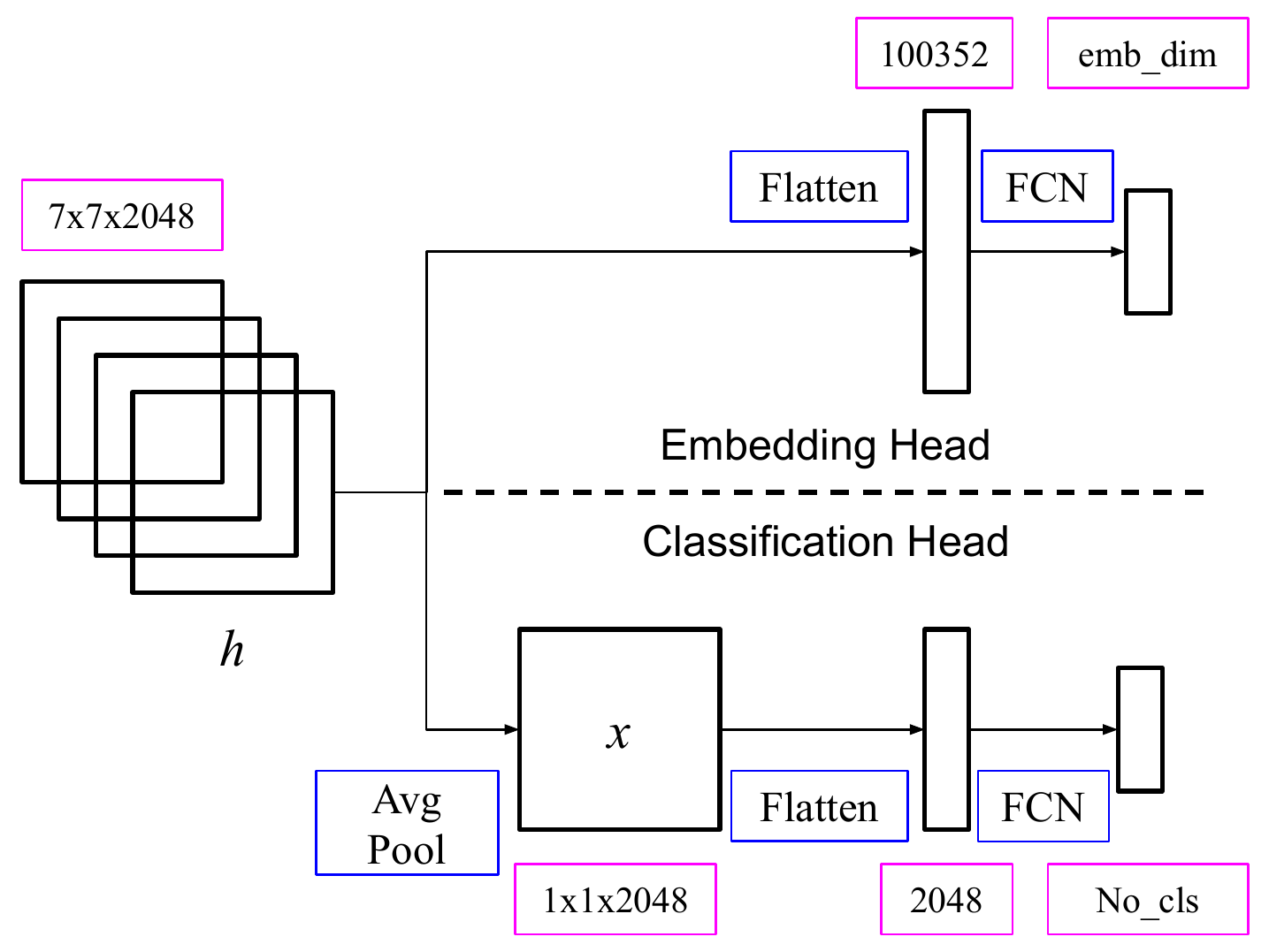}
	\caption{Our proposed two-head architecture. The last convolutional feature map ($h$) supports both embedding and classification heads. Operations and dimensions are highlighted with blue and pink colors, respectively. ResNet-50 dimensions used for illustration.}
	\label{fig:two_head_arch_details}
\end{figure}

Unlike VGG~\cite{simonyan2014very}, recent architectures~\cite{he2016deep,szegedy2016rethinking,huang2017densely} employ a convolutional layer before the classification head. To generate logits, the classification head pools the convolutional layer features, flatten them, then utilize a customizable fully connected layer to support various numbers of classes. Similarly, we integrate triplet loss to regularize embedding as shown in Figure~\ref{fig:two_head_arch_details}. Before pooling, we flatten the convolutional layer features then apply another fully connected layer $W_{\text{emb}}$ to generate embeddings as illustrated in equation~\ref{eq:w_emd}. 
\begin{align}
	\text{Logits} &= W_{\text{logits}} * \text{flatten}(x) \\
	\text{Embedding} &= W_{\text{emb}} * \text{flatten}(h) \label{eq:w_emd},
\end{align}
where $x= \text{pool}(h)$. Orderless pooling, like averaging, disregard spatial information. Thus, a fully connected layer $W_{\text{emb}}$ applied on $h$ has a better representation power.  The final embedding is normalized to the unit-circle and the square Euclidean distance metric is employed. During inference, the two-head architecture enables both classification and retrieval with negligible overhead.


\section{Experiments}

\subsection{Evaluation on FGVR}

\newcommand*\rot{\rotatebox{90}}

\begin{table}[t]
	\scriptsize
	\begin{center}
		\begin{tabular}{@{}lccccc@{}}
			\hline
			& \rot{Flowers-102~\cite{nilsback2008automated}} &  \rot{Aircrafts~\cite{maji2013fine}} &  \rot{NABirds~\cite{van2015building}}  &  \rot{Stanford Cars~\cite{krause20133d}} &  \rot{Stanford Dogs~\cite{khosla2011novel}} \\ 		\hline
			Num Classes           & 102         &  100&  550 & 196 &  120\\ 		
			Avg samples Per Class & 10          &  100 &  43.5 &  41.55  & 100\\ 		
			Train Size            & 1020        &  3334 & 23929	  & 8144 & 12000\\ 		
			Val Size              & 1020        & 3333 & N\slash A  & N\slash A & N\slash A \\ 	
			Test Size             & 6149        &  3333 & 24633  &	8041 & 8580	\\ \hline
			Total Size             & 8189        &  10000 & 48562  &	16185 & 20580	\\ \hline
		\end{tabular}
	\end{center}
	\caption{Statistics of five FGVR datasets and their corresponding train, validation and test splits.}
	\label{tbl:fgvr}
\end{table}

\newcommand{\imgsize}{0.09}

\noindent\textbf{\underline{Datasets:}} We evaluate our approach on five FGVR datasets. These datasets comprise both make/model classification and wildlife species. The \textbf{Aircrafts} dataset contains 10,000 images of aircraft spanning 100 aircraft-models. The finer level differences between models makes visual recognition challenging. The \textbf{NABirds} dataset contains 48,562 images across 550 visual categories of North American birds. The \textbf{Flower-102} dataset contains 8189 images across 102 classes. The \textbf{Stanford Cars} dataset contains 16185 images across 196 car classes that represent variations in car make, model, and year. Finally, the \textbf{Stanford Dogs} dataset has 20,580 images across 120 breeds of dogs. These datasets provide challenges in terms of large intra-class but small inter-class variations. Table~\ref{tbl:fgvr} summarizes the datasets' size, number of classes and splits. 

\newcommand\sbullet[1][.5]{\mathbin{\vcenter{\hbox{\scalebox{#1}{$\bullet$}}}}}
\noindent\textbf{\underline{Baselines:}} We evaluate our approach against two baselines: (1) Single head softmax; (2) Two-head leveraging center loss~\cite{wen2016discriminative} with it's proposed hyper-parameters  $\lambda=0.003$ and $\alpha=0.5$. We found Magnet loss~\cite{rippel2015metric} implementation computationally expensive. It applies k-means to cluster all training samples after each epoch, \ie, $O(N^2)$ where $N$ is the train split size. For triplet loss, both hard~\cite{hermans2017defense} and semi-hard~\cite{schroff2015facenet} sampling variants are evaluated. By default, our hyper-parameter $\lambda=1$ and embedding normalized to the unit circle with dimensionality $d_{\text{emb}}=256$. With triplet hard sampling, a soft margin between classes is imposed by the softplus function $\ln(1+\exp( \sbullet[.75] ))$. It is similar to the hinge function $\max(\sbullet[.75] ,0)$ but it decays exponentially instead of a hard cut-off. With triplet semi-hard sampling, we employ the hard margin $m=0.2$ as proposed by~\cite{schroff2015facenet}

All  experiments are conducted on Titan Xp 12GB GPU with batch-size $b=32$. All networks are initialized with ImageNet weights, and then fine-tuned. Momentum optimizer is utilized with momentum $0.9$ and a polynomial decaying learning rate  $lr=0.01$. We quantitatively evaluate our approach on three architectures: (1) ResNet-50~\cite{he2016deep} and (2) DenseNet-161~\cite{huang2017densely} both trained for 40K iterations, and (3) Inception-V4~\cite{szegedy2017inception} trained for 80K iterations.  While early stopping is a valid regularization form to avoid a fixed number of training iteration, not all datasets provide a validation split as illustrated in table~\ref{tbl:fgvr}. The chosen number of training iterations achieve comparable results with recent FGVR softmax baselines~\cite{lin2017improved,krause2016unreasonable,dubey2018pairwise}.
 

To evaluate our approach, our training batches contain both positive and negative samples. We follow the batch construction procedure proposed by Hermans \etal~\cite{hermans2017defense}. A class is uniformly sampled then $K = 4$ sample images, with resolution $224\times 224$, are randomly drawn. Training images are augmented online with random cropping and horizontal flipping. This process iterates until a batch is complete.  Table~\ref{tbl:fgvr_quan_eval} presents our fine-tuning quantitative evaluation on the five datasets. Our two-head architecture with hard triplet loss achieves large steady (1-4\%) improvement on ResNet-50. Similar trend appears with Inception-V4 but suffers an interesting fluctuation between hard and semi-hard triplet loss. Section~\ref{subsec:retrieval} reflects on this phenomena through a quantitative embedding analysis. Vanilla DenseNet-161 achieves comparable state-of-the-art results on all FGVR datasets, yet triplet loss regularizer maintains a steady trend of performance improvement. 

	Center loss achieves an inferior classification performance especially on the Dogs dataset -- a lag $\approx 4\%$ behind vanilla softmax on Inception-V4 and DenseNet-161. The single mode embedding assumption is valid for face recognition~\cite{wen2016discriminative} and vehicle re-identification~\cite{liu2016deep} because different images for the same identify belong to a single cluster. However, when working with categories of high intra-class variations, this assumption degenerate the feature embedding quality. Our feature embedding evaluation (Sec ~\ref{subsec:retrieval}) highlights the consequence of using a single mode/cluster, for general classification problems, in terms of feature embedding instability or collapse.  
 
Our simple but vital integration into standard architectures distance our approach from similar softmax+clustering formulations. In addition, all recent convolutional architectures share similar ending structure; the last convolutional layer is followed by an average pooling, and then a single fully connected layer. Thus, apart from the studied architectures, our secondary embedding head proposal can be applied to other architectures, \eg, MobileNet~\cite{howard2017mobilenets}.

\begin{table}
		\scriptsize
		\centering
		\begin{tabular}{@{}lccccc@{}}
		\hline
		Database          & Cars & Flowers & Dogs  & Aircrafts & Birds \\ \hline
		\multicolumn{6}{c}{ResNet-50} \\
		\hline
		Softmax         &   85.85  & 85.68 & 69.76 & 83.22 & 64.23 \\
		\textbf{Two-Head (Center)} &  88.23  & 85.00 & 70.45 & 84.48  & 65.5 \\ 
		\textbf{Two-Head (Semi)} &  88.22  & 85.52 & 70.69 & 85.08 & 65.20 \\ 
		\textbf{Two-Head (Hard)} &  \textbf{89.44} &  \textbf{86.61} & \textbf{72.70} & \textbf{87.33} & \textbf{66.19} \\ \hline
		
		\multicolumn{6}{c}{Inception-V4} \\
		\hline
		Softmax         &   88.42  & 88.22 & 77.20 & 86.76 & 74.90 \\
		\textbf{Two-Head (Center)} &  89.50  & 88.35 & 70.83  & 87.78  & 76.86 \\ 
		\textbf{Two-Head (Semi)} &  \textbf{89.72}  & 88.69 & \textbf{77.71} & 88.59 & \textbf{76.99} \\ 
		\textbf{Two-Head (Hard)} &  89.06 &  \textbf{90.66} & 75.97 & \textbf{89.04} & 76.57 \\ \hline
		\multicolumn{6}{c}{DenseNet-161} \\
		\hline
		Softmax         &   91.64  & 92.56 & \textbf{81.58} & 89.13 & 78.69 \\
		\textbf{Two-Head (Center)} &  89.08 &  92.58 & 77.02 & \textbf{89.97} & 79.05 \\
		\textbf{Two-Head (Semi)} &  \textbf{92.36}  & \textbf{93.65} & 80.89 & 89.64 & \textbf{79.57} \\ 
		\textbf{Two-Head (Hard)} &  92.41 &  93.25 & 81.16 & 89.34 & 79.47 \\
		\hline
	\end{tabular}	
\caption{Quantitative evaluation on the five FGVR datasets using ResNet-50, Inception-V4, and DenseNet-161.}
\label{tbl:fgvr_quan_eval}
\end{table}

\subsection{Task Generalization}
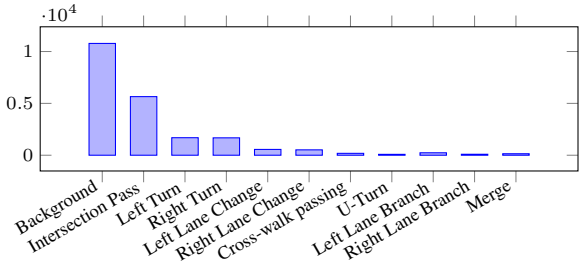
\begin{figure}
	\scriptsize
	\begin{tikzpicture} \begin{axis}[ ybar,width=0.5\textwidth, height=3.5cm, enlargelimits=0.15, symbolic x coords={Background,Intersection Pass,Left Turn,Right Turn,Left Lane Change,Right Lane Change,Cross-walk passing,U-Turn,Left Lane Branch,Right Lane Branch,Merge}, xtick=data, nodes near coords align={vertical}, x tick label style={rotate=30,anchor=east},] 
	\addplot coordinates {
		(Background,10781)
		(Intersection Pass,5651) 
		(Left Turn,1689) 
		(Right Turn,1677) 
		(Left Lane Change,560) (Right Lane Change,518) (Cross-walk passing,182) (U-Turn,85) (Left Lane Branch,235) (Right Lane Branch,93) (Merge,143)};  \end{axis}
	\end{tikzpicture}
	\caption{Honda driving dataset long tail class distribution}
	\label{fig:HDD_dist}
\end{figure}

For further evaluation, we leverage the Honda Research Institute Driving Dataset (HDD)~\cite{RamanishkaCVPR2018} for action recognition. HDD is an ego-motion video dataset for driver behavior understanding and causal reasoning. It contains 10,833 events spanning eleven event classes. Moreover, the HDD event class distribution is long-tailed which poses an imbalance data challenge. Figure~\ref{fig:HDD_dist} shows the eleven event classes with their distributions. To reduce video frames' redundancy, three frames are sampled per second, and events shorter than 2 seconds are omitted.

\begin{figure}
	\centering
	\begin{subfigure}{1.0\linewidth}
		\centering
		
		\includegraphics[width=0.20\linewidth]{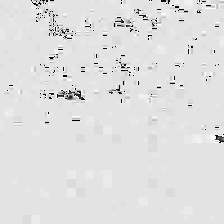}
		\includegraphics[width=0.20\linewidth]{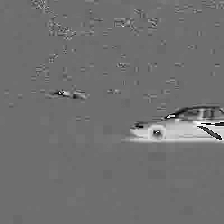}	
	\end{subfigure}
	\begin{subfigure}{1.0\linewidth}
		\centering
		\includegraphics[width=0.20\linewidth]{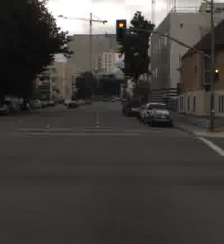}
		\includegraphics[width=0.20\linewidth]{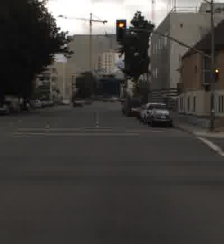}
		\includegraphics[width=0.20\linewidth]{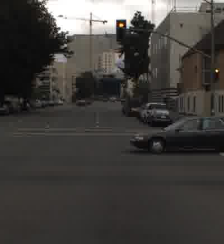}
	\end{subfigure}
	\caption{Stack of difference motion encoding. Instead of six frames, three are used for visualization purpose. The first row shows a stack of two difference frames constructed by subtracting consecutive pairs of grayscale frames in the second row. These images are best viewed in color/screen.}
	\label{fig:sod}
\end{figure}
To leverage standard architecture for action recognition, stack of difference (SOD) motion encoding proposed by Fernando et al.~\cite{fernando2017self}  is adopted. While better motion encoding like optical-flow exists, the SOD is utilized for its simplicity and ability to achieve competitive results~\cite{fernando2017self,taha2018two}. Given a sequence of frames representing an event, six consecutive frames spanning 2 seconds are randomly sampled. They are converted to grayscale, and then every consecutive pair is subtracted to create a stack of difference $\in Z^{W\times H\times 5}$ as depicted in Figure~\ref{fig:sod}. Standard architectures are easily adapted to this input representation  by treating the SOD input as a five-channel image instead of three. 

Unlike FGVR input $\in[0,255]$, SOD $\in [-255,255]$. Thus, a ResNet-50~\cite{he2016deep} architecture initialized with random weights is employed. It is trained for 10K iterations with  $\lambda=1$  and a polynomial decaying learning rate $lr=0.01$. Batch sizes 33 and 63 are used to compare the vanilla softmax against our approach. To highlight performance on minority classes, both micro and macro average accuracies are reported in Table~\ref{tbl:honda_acc}. Macro-average computes the metric for each class independently before taking the average. Micro-average is the traditional mean for all samples. Macro-average treats all classes equally while micro-averaging favors majority classes. Table~\ref{tbl:hdd_macro_eval} highlights the efficiency of our approach on minority classes.

\begin{table}[]
	\scriptsize
	\centering
	\begin{tabular}{@{}lcc@{}}
		\hline
		& Micro Acc & Macro  Acc \\ \hline
		Softmax  ($b=33$)        & 84.43    & 47.66                      \\ 
				Two-head (Semi) ($b=33$) &  \textbf{84.93}    & \textbf{53.70}                     \\ \hline 
		Softmax ($b=63$)        & 84.45    & 46.53                  \\ 
		Two-head (Semi) ($b=63$) & \textbf{84.85}    & \textbf{54.08}                      \\ \hline
	\end{tabular}
\caption{Action recognition quantitative evaluation on the Honda dataset. $b$ indicates the batch-size used. Macro average accuracy highlights performance on minority classes.}
\label{tbl:honda_acc}
\end{table}




\newcommand{\graycellcolor}{gray!50}
\begin{table}
	\scriptsize
	\centering
	\begin{tabular}{@{}lcc|cc@{}}
		\hline
		& Softmax & \textbf{Two-Head} & Softmax & \textbf{Two-Head} \\ \hline
		Event &\multicolumn{2}{c|}{Batch-size 33}& \multicolumn{2}{c}{Batch-size 63} \\ \hline 
		Background				& \cellcolor{\graycellcolor}\textbf{96.28}&95.29&\cellcolor{\graycellcolor}\textbf{97.32}& 96.28 \\ 
		Intersection Passing &74.61&\cellcolor{\graycellcolor}\textbf{75.86}&74.26& \cellcolor{\graycellcolor}\textbf{74.68} \\ 
		Left Turn					&\cellcolor{\graycellcolor}\textbf{85.49}&84.87&85.18& \cellcolor{\graycellcolor}\textbf{86.11} \\ 
		Right Turn 					&\cellcolor{\graycellcolor}\textbf{88.47}&87.22&\cellcolor{\graycellcolor}\textbf{86.91}& 86.60 \\ 
		Left Lane Change 	&59.40&\cellcolor{\graycellcolor}\textbf{66.33}&55.44& \cellcolor{\graycellcolor}\textbf{62.37} \\ 
		Right Lane Change 	&44.79& \cellcolor{\graycellcolor}\textbf{61.45}&40.62& \cellcolor{\graycellcolor}\textbf{51.04} \\ 
		Cross-walk Passing &\cellcolor{\graycellcolor}\textbf{18.18}&\cellcolor{\graycellcolor}\textbf{18.18}&\cellcolor{\graycellcolor}\textbf{12.12}&\cellcolor{\graycellcolor}\textbf{12.12}  \\ 
		U-Turn 						&0.00&\cellcolor{\graycellcolor}\textbf{11.76}&0.00& \cellcolor{\graycellcolor}\textbf{23.52} \\ 
		Left Lane Branch 		&53.84&\cellcolor{\graycellcolor}\textbf{64.10}&41.02&\cellcolor{\graycellcolor} \textbf{64.10} \\ 
		Right Lane Branch	&0.00&\cellcolor{\graycellcolor}\textbf{6.24}&12.49&\cellcolor{\graycellcolor} \textbf{18.74} \\ 
		Merge 						&3.22&\cellcolor{\graycellcolor}\textbf{19.35}&6.45& \cellcolor{\graycellcolor}\textbf{19.35} \\ \hline
		Macro Accuracy & 47.66 & \textbf{53.70} &46.53& \textbf{54.08} \\\hline
	\end{tabular}
\caption{Detailed evaluation on the Honda driving dataset. Our two-head architecture using semi-hard triplet loss achieves better performance on minority classes.}
\label{tbl:hdd_macro_eval}
\end{table}


\subsection{Retrieval Evaluation on FGVR}\label{subsec:retrieval}

In the two-head architecture, the secondary embedding head brings values like an enhanced feature embedding, nearest neighbor retrieval and interpretability. Following  Song \etal~\cite{oh2016deep}, we evaluate the quality of feature embedding using Recall@K metric on the test split. We also leverage the Normalized Mutual Info (NMI) score to evaluate the quality of cluster alignments. $\text{NMI}=\frac { I(\Omega ,C) }{ \sqrt { H(\Omega )H(C) }  } ,$ where $\Omega =\{\omega_1,..,\omega_n\}$ is the ground-truth clustering while $C=\{c_1,...c_n\}$ is a clustering assignment for the learned embedding. $I(\sbullet[0.5],\sbullet[0.5])$ and $H(\sbullet[0.5])$ denotes mutual information and entropy respectively. We use K-means to compute $C$.


\begin{table}[t]
	\scriptsize
	\centering
	\begin{tabular}{@{}llccccc@{}}
		&      & NMI   & R@1     & R@4   & R@8   & R@16  \\ \hline
		\multirow{3}{*}{Car - ResNet} & CNTR & 0.549  & 67.73 & 75.36 & 81.91 & 87.28 \\
		& SEMI & 0.879 & 89.45 & 93.14 & 95.24 & 96.62 \\
		& HARD & \textbf{0.900} & \textbf{91.95} & \textbf{94.22} & \textbf{95.70}  & \textbf{96.78} \\ \hline
		
		\multirow{3}{*}{Flowers - ResNet} & CNTR & 0.723 & 74.53 & 86.78 & 90.94 & 94.06 \\
		& SEMI & 0.822 & 87.56 & \textbf{94.29} & \textbf{96.39} & \textbf{97.89} \\
		& HARD & \textbf{0.856} & \textbf{90.40} & 94.00 & 94.84 & 95.64 \\ \hline
		
		\multirow{3}{*}{Dogs - ResNet} & CNTR & 0.419 & 30.41 & 40.69 & 63.96 & 75.14 \\
		& SEMI & 0.708 & 60.70 & 79.55 & 85.84 & 90.15 \\
		& HARD & \textbf{0.740} & \textbf{64.01} & \textbf{81.60} & \textbf{86.41} & \textbf{89.97} \\ \hline

		\multirow{3}{*}{Aircrafts - ResNet} & CNTR & 0.645 & 64.36 & 80.32 & 85.57 & 89.41 \\
		& SEMI & 0.846 & 82.15 & 90.01 & 92.38 & \textbf{94.45} \\
		& HARD & \textbf{0.879} & \textbf{85.84} & \textbf{91.63} & \textbf{92.89} & 93.94 \\ \hline
		
		\multirow{3}{*}{NABirds - ResNet} & CNTR & 0.517 & 32.16 & 50.89 & 60.03 & 68.70 \\
		& SEMI & 0.749 & 56.30 & 76.08 & 82.99 & \textbf{88.30}  \\
		& HARD & \textbf{0.769} & \textbf{59.09} & \textbf{77.35} & \textbf{83.49} & 88.12 \\ \hline \hline
		
		\multirow{3}{*}{Cars - Inc-V4} & CNTR & 0.120 & 2.98  & 5.96  & 8.84  & 13.87 \\
		& SEMI & \textbf{0.880}  & \textbf{85.45} & \textbf{93.56} & \textbf{95.66} & \textbf{97.15} \\
		& HARD & 0.652 & 46.97 & 71.14 & 80.87 & 87.90 \\ \hline
		
		\multirow{3}{*}{Flowers - Inc-V4} & CNTR & 0.183 & 9.01  & 11.97 & 13.82 & 16.13 \\
		& SEMI & 0.828 & 88.70 & 94.70 & 96.47 & \textbf{97.89} \\
		& HARD & \textbf{0.885} & \textbf{93.66} & \textbf{96.13} & \textbf{96.96} & 97.59 \\ \hline
		
		\multirow{3}{*}{Dogs - Inc-V4} & CNTR & 0.726 & 65.47 & 76.62 & 79.01 & 81.04 \\
		& SEMI & \textbf{0.760} & \textbf{68.48} & \textbf{85.10} & \textbf{90.26} & \textbf{93.83} \\
		& HARD & 0.458 & 19.52 & 41.41 & 55.63 & 70.63 \\ \hline

		\multirow{3}{*}{Aircrafts - Inc-V4} & CNTR & 0.333 & 27.21 & 36.75 & 42.81 & 49.62 \\
		& SEMI & 0.872 & 86.53 & 92.35 & \textbf{93.88} & \textbf{95.08} \\
		& HARD & \textbf{0.887} & \textbf{87.79} & \textbf{92.47} & 93.67 & 94.42 \\ \hline
		
		\multirow{3}{*}{NABirds - Inc-V4} & CNTR & 0.209 & 3.77  & 6.26  & 8.29  & 11.50 \\
		& SEMI & \textbf{0.808} & \textbf{67.30} & \textbf{83.81} & \textbf{88.96} & \textbf{92.79} \\
		& HARD & 0.503 & 15.92 & 31.84 & 42.66 & 54.64 \\ \hline \hline
		
		\multirow{3}{*}{Cars - Dense} & CNTR & \textbf{0.914} & 88.93 & 93.97 & 95.01 & 95.65 \\
		& SEMI & 0.905 & 88.77 & \textbf{95.72} & \textbf{97.08} & \textbf{98.30} \\
		& HARD & 0.913 & \textbf{89.40} & 95.57 & 96.99 & 98.15 \\ \hline
		
		\multirow{3}{*}{Flowers - Dense} & CNTR & \textbf{0.910} & \textbf{95.23} & 97.19 & 97.61 & 98.13 \\
		& SEMI & 0.869 & 94.52 & \textbf{97.90} & \textbf{98.68} & \textbf{99.14} \\
		& HARD & 0.898 & 87.73 & 91.87 & 92.32 & 92.65 \\ \hline
		
		\multirow{3}{*}{Dogs - Dense} & CNTR & 0.795 & 72.03 & 84.11 & 86.55 & 88.39 \\
		& SEMI & 0.802 & 73.33 & 88.24 & 92.21 & \textbf{95.02} \\
		& HARD & \textbf{0.807} & \textbf{73.99} & \textbf{88.66} & \textbf{92.44} & 94.99 \\ \hline
		
		\multirow{3}{*}{Aircrafts - Dense} & CNTR & \textbf{0.898} & 87.73 & 91.87 & 92.32 & 92.65 \\
		& SEMI & 0.883 & 86.98 & 93.49 & 95.11 & \textbf{96.28} \\
		& HARD & 0.889 & \textbf{87.82} & \textbf{94.27} & \textbf{95.38} & 96.07	\\ \hline
		
		\multirow{3}{*}{NABirds - Dense} & CNTR & \textbf{0.847} & \textbf{76.90} & 85.37 & 88.03 & 90.57 \\
		& SEMI & 0.829 & 72.09 & 86.90 & \textbf{91.24} & 94.35 \\
		& HARD & 0.829 & 72.02 & \textbf{87.11} & 91.61 & \textbf{94.70} \\ \hline \hline
	\end{tabular}
	\caption{Detailed feature embedding quantitative analysis across the five datasets using ResNet-50, Inception-V4 and DenseNet-161. Triplet with hard mining achieves superior embedding with ResNet-50 trained for 40K iterations. Semi-hard triplet is competitive and stable with Inception-V4 trained for 80K iterations. Center loss learns an inferior embedding while suffering the highest instability.}
	\label{tbl:embedding_qual}
\end{table}

Table~\ref{tbl:embedding_qual} presents a detailed feature embedding quantitative analysis. Triplet loss with hard-mining consistently learns the best embedding on ResNet-50. However, semi-hard sampling, on Inception-V4 and DenseNet, is stabler.  Despite having an explicit rebelling force pushing negative samples away from their anchors, hard triplet mining can in practice lead to bad local minima (as can be seen in inception-V4). It can result in a collapsed mode $(\ie, f(x) = 0)$~\cite{schroff2015facenet}. Center loss suffers the same model collapse problem. It is a more vulnerable variant of hard-triplet loss, i.e., missing the repelling force. It learns an inferior embedding while suffering the highest instability. It often degenerates with Inception-V4.  These conclusions follow Schroff \etal~\cite{schroff2015facenet} semi-hard mining findings.



Table~\ref{tbl:retfgvr} compares classification and retrieval performance quantitatively. The reported classification accuracy provides an upper bound for retrieval. Retrieval and classification top 1 accuracies are comparable. Recall@4 is superior to the classification top 1 on all datasets. Figure~\ref{fig:cross_qual} presents a qualitative retrieval evaluation across center loss, triplet semi-hard, and triplet hard regularizers. 




\begin{table}[t]
	\scriptsize
	\centering
	\begin{center}
		\begin{tabular}{@{}lccccc@{}}
			\hline
			& Cars & Flowers-102 &  Dogs & Aircrafts &  NABirds  \\ 		\hline
			& \multicolumn{5}{c}{ResNet-50} \\
			Classification Top 1           & 89.44         &  86.61 & 72.70 & 87.33& 66.19  \\  
			Retrieval Top 1 & 91.95         &  90.40 &  64.01   &85.84 & 59.09 \\ 	
			Retrieval Top 4 & 94.22         &  94.00 &  81.60   & 91.63 & 77.35 \\ 		\hline
			
			& \multicolumn{5}{c}{Inception-V4} \\
			Classification Top 1           & 89.72         &  90.66 & 77.71 & 89.04 & 76.99  \\  
			Retrieval Top 1 & 85.45         &  93.66 &  68.48   & 87.79 & 67.30 \\ 	
			Retrieval Top 4 & 93.56         &  96.13 &  85.10  & 92.47 & 83.81 \\ 		\hline

			& \multicolumn{5}{c}{DenseNet-161} \\
			Classification Top 1           & 92.36        & 93.65 & 81.58 & 89.97 & 76.57 \\  
			Retrieval Top 1 & 89.40         &  95.23 &  73.99   & 87.82 & 76.90 \\ 	
			Retrieval Top 4 & 95.72         &  97.90 &  88.66  & 94.27 & 87.11 \\ 		\hline
			
		\end{tabular}
	\end{center}
	\caption{Comparative quantitative evaluation between retrieval  and classification as an upper bound. Both retrieval and classification accuracies are comparable. Retrieval top 4 is superior to classification top 1.}
	\label{tbl:retfgvr}
\end{table}

\newcommand{\CrossQualImgSize}{0.075}
\renewcommand{\arraystretch}{0.5}
\newcommand{\CrossQualVCenter}{0.0725} 
\setlength{\fboxsep}{0pt}%
\setlength{\fboxrule}{1pt}%
\begin{figure*}
	\centering
	\setlength\tabcolsep{0.5pt} 
	\begin{tabular}{cccccccccccc}
		Query $\downarrow$&
		\fcolorbox{green}{white}{\includegraphics[width=\CrossQualImgSize\textwidth,height=\CrossQualImgSize\textwidth]{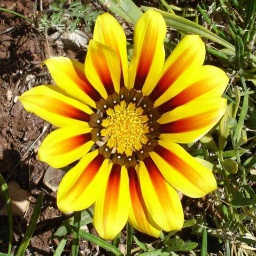}} & \fcolorbox{red}{white}{\includegraphics[width=\CrossQualImgSize\textwidth,height=\CrossQualImgSize\textwidth]{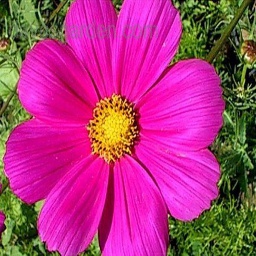}} & 
		\fcolorbox{red}{white}{\includegraphics[width=\CrossQualImgSize\textwidth,height=\CrossQualImgSize\textwidth]{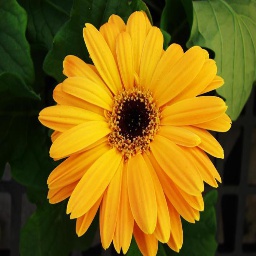}} &
		
 Query $\downarrow$&

\fcolorbox{green}{white}{\includegraphics[width=\CrossQualImgSize\textwidth,height=\CrossQualImgSize\textwidth]{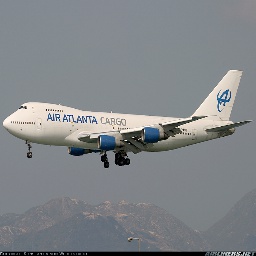}} & \fcolorbox{red}{white}{\includegraphics[width=\CrossQualImgSize\textwidth,height=\CrossQualImgSize\textwidth]{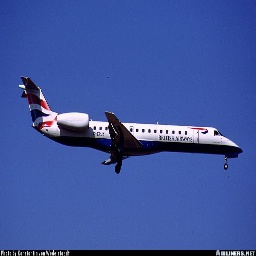}} & 
\fcolorbox{red}{white}{\includegraphics[width=\CrossQualImgSize\textwidth,height=\CrossQualImgSize\textwidth]{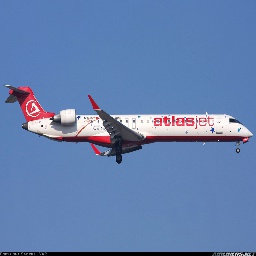}} &

 Query $\downarrow$&

\fcolorbox{red}{white}{\includegraphics[width=\CrossQualImgSize\textwidth,height=\CrossQualImgSize\textwidth]{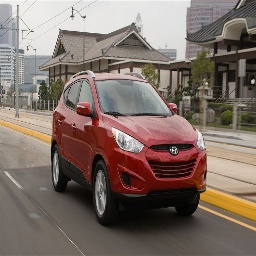}} & \fcolorbox{red}{white}{\includegraphics[width=\CrossQualImgSize\textwidth,height=\CrossQualImgSize\textwidth]{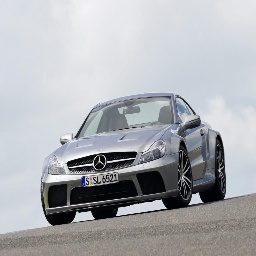}} & 
\fcolorbox{red}{white}{\includegraphics[width=\CrossQualImgSize\textwidth,height=\CrossQualImgSize\textwidth]{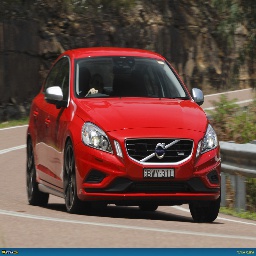}} \\

		\includegraphics[width=\CrossQualImgSize\textwidth,height=\CrossQualImgSize\textwidth]{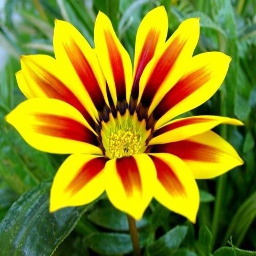}&
\fcolorbox{green}{white}{\includegraphics[width=\CrossQualImgSize\textwidth,height=\CrossQualImgSize\textwidth]{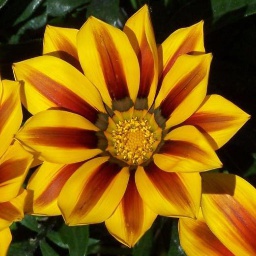}} & \fcolorbox{green}{white}{\includegraphics[width=\CrossQualImgSize\textwidth,height=\CrossQualImgSize\textwidth]{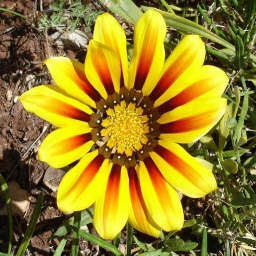}} & 
\fcolorbox{green}{white}{\includegraphics[width=\CrossQualImgSize\textwidth,height=\CrossQualImgSize\textwidth]{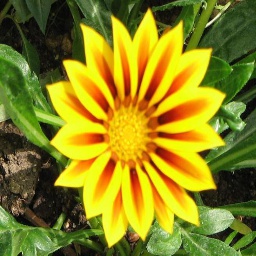}} &

\includegraphics[width=\CrossQualImgSize\textwidth,height=\CrossQualImgSize\textwidth]{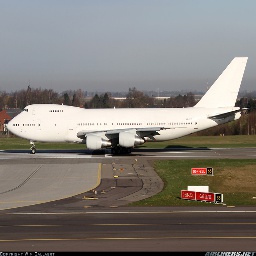}&
\fcolorbox{green}{white}{\includegraphics[width=\CrossQualImgSize\textwidth,height=\CrossQualImgSize\textwidth]{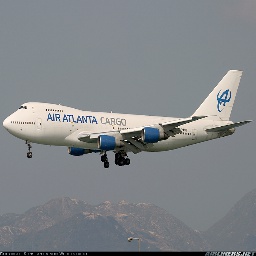}} & \fcolorbox{red}{white}{\includegraphics[width=\CrossQualImgSize\textwidth,height=\CrossQualImgSize\textwidth]{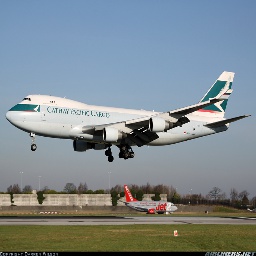}} & 
\fcolorbox{red}{white}{\includegraphics[width=\CrossQualImgSize\textwidth,height=\CrossQualImgSize\textwidth]{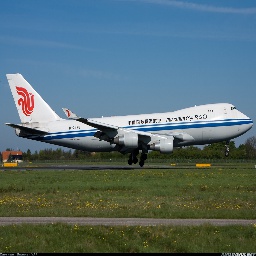}} &

\includegraphics[width=\CrossQualImgSize\textwidth,height=\CrossQualImgSize\textwidth]{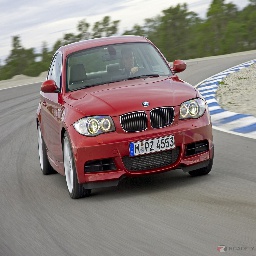}&
\fcolorbox{green}{white}{\includegraphics[width=\CrossQualImgSize\textwidth,height=\CrossQualImgSize\textwidth]{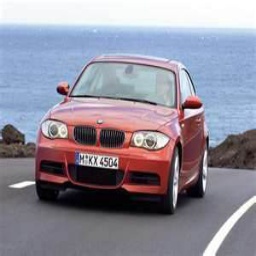}} & \fcolorbox{green}{white}{\includegraphics[width=\CrossQualImgSize\textwidth,height=\CrossQualImgSize\textwidth]{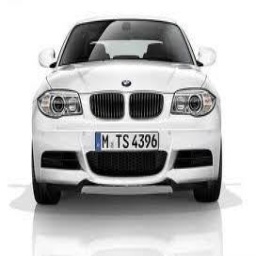}} & 
\fcolorbox{green}{white}{\includegraphics[width=\CrossQualImgSize\textwidth,height=\CrossQualImgSize\textwidth]{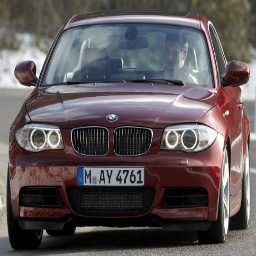}} \\

&
\fcolorbox{green}{white}{\includegraphics[width=\CrossQualImgSize\textwidth,height=\CrossQualImgSize\textwidth]{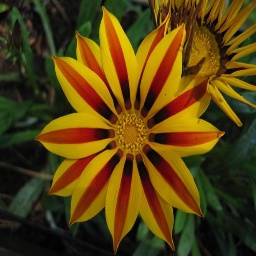}}& \fcolorbox{green}{white}{\includegraphics[width=\CrossQualImgSize\textwidth,height=\CrossQualImgSize\textwidth]{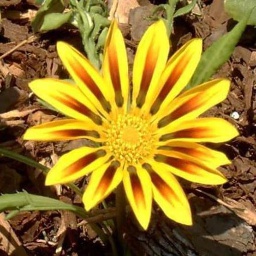}} & 
\fcolorbox{green}{white}{\includegraphics[width=\CrossQualImgSize\textwidth,height=\CrossQualImgSize\textwidth]{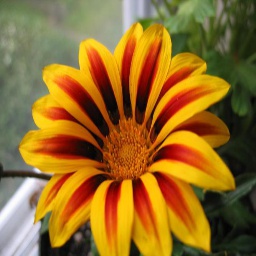}} &

&
\fcolorbox{green}{white}{\includegraphics[width=\CrossQualImgSize\textwidth,height=\CrossQualImgSize\textwidth]{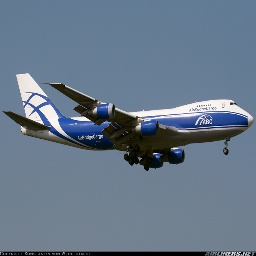}} & \fcolorbox{green}{white}{\includegraphics[width=\CrossQualImgSize\textwidth,height=\CrossQualImgSize\textwidth]{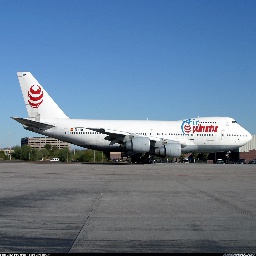}} & 
\fcolorbox{green}{white}{\includegraphics[width=\CrossQualImgSize\textwidth,height=\CrossQualImgSize\textwidth]{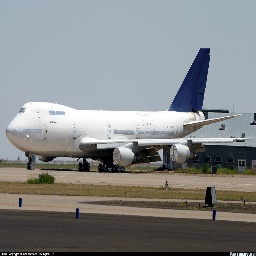}} &

&
\fcolorbox{green}{white}{\includegraphics[width=\CrossQualImgSize\textwidth,height=\CrossQualImgSize\textwidth]{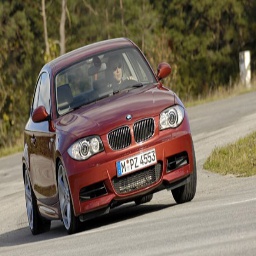}} & \fcolorbox{green}{white}{\includegraphics[width=\CrossQualImgSize\textwidth,height=\CrossQualImgSize\textwidth]{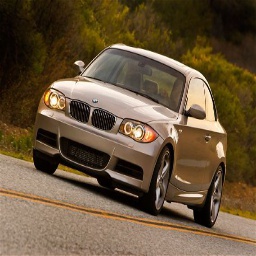}} & 
\fcolorbox{red}{white}{\includegraphics[width=\CrossQualImgSize\textwidth,height=\CrossQualImgSize\textwidth]{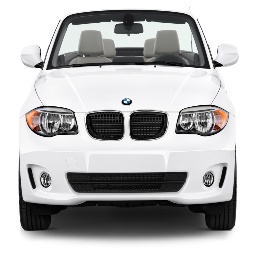}} \\

	\end{tabular}

\caption{Retrieval qualitative evaluation on three FGVR datasets: Flowers-102, Aircrafts and Cars. Given a query image, the three nearest neighbors are depicted. The three consecutive rows show search results using center loss, semi-hard and hard triplet regularizers. Green and red outlines denote match and mismatch between the query and it's result respectively.}
\label{fig:cross_qual}
\end{figure*}

\begin{figure}[t]
	\centering
	\begin{tabular}{cc|cc|cc}
		
		
		
		\includegraphics[width=\CrossQualImgSize\textwidth,height=\CrossQualImgSize\textwidth]{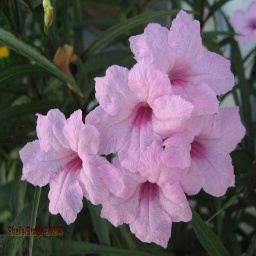}\includegraphics[width=\CrossQualImgSize\textwidth,height=\CrossQualImgSize\textwidth]{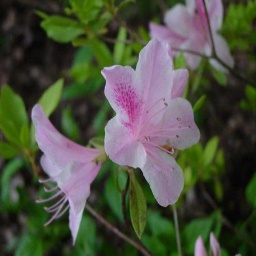}
		\includegraphics[width=\CrossQualImgSize\textwidth,height=\CrossQualImgSize\textwidth]{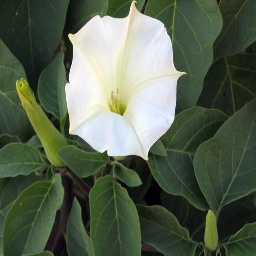}\includegraphics[width=\CrossQualImgSize\textwidth,height=\CrossQualImgSize\textwidth]{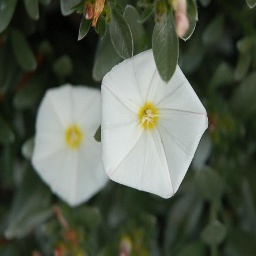}
		\includegraphics[width=\CrossQualImgSize\textwidth,height=\CrossQualImgSize\textwidth]{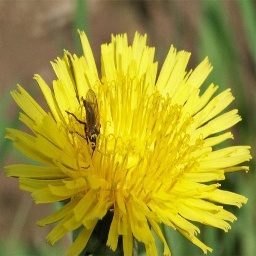}\includegraphics[width=\CrossQualImgSize\textwidth,height=\CrossQualImgSize\textwidth]{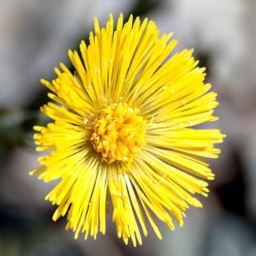}
		\\
		
		\includegraphics[width=\CrossQualImgSize\textwidth,height=\CrossQualImgSize\textwidth]{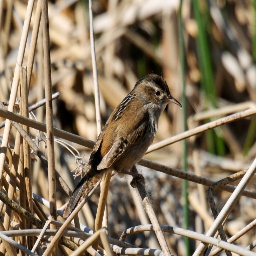}\includegraphics[width=\CrossQualImgSize\textwidth,height=\CrossQualImgSize\textwidth]{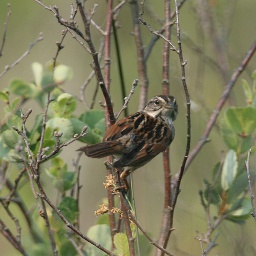}
		\includegraphics[width=\CrossQualImgSize\textwidth,height=\CrossQualImgSize\textwidth]{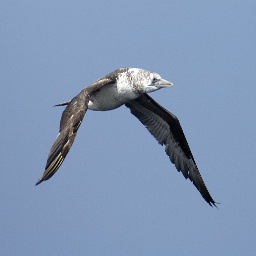}\includegraphics[width=\CrossQualImgSize\textwidth,height=\CrossQualImgSize\textwidth]{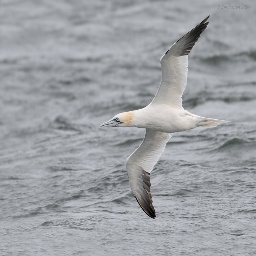}
		\includegraphics[width=\CrossQualImgSize\textwidth,height=\CrossQualImgSize\textwidth]{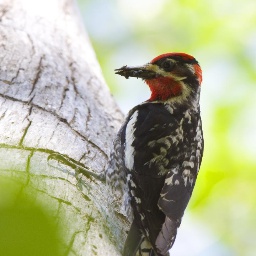}\includegraphics[width=\CrossQualImgSize\textwidth,height=\CrossQualImgSize\textwidth]{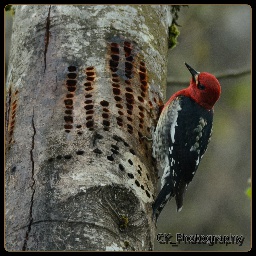}
		\\
		
		
		
		
		
		\includegraphics[width=\CrossQualImgSize\textwidth,height=\CrossQualImgSize\textwidth]{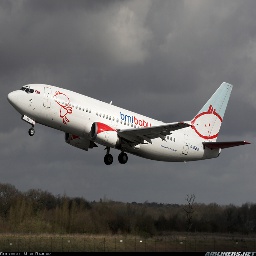}\includegraphics[width=\CrossQualImgSize\textwidth,height=\CrossQualImgSize\textwidth]{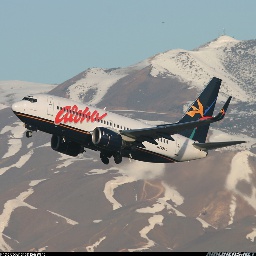}
		\includegraphics[width=\CrossQualImgSize\textwidth,height=\CrossQualImgSize\textwidth]{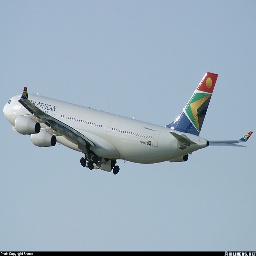}\includegraphics[width=\CrossQualImgSize\textwidth,height=\CrossQualImgSize\textwidth]{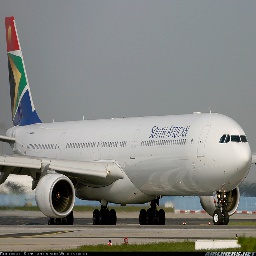}
		\includegraphics[width=\CrossQualImgSize\textwidth,height=\CrossQualImgSize\textwidth]{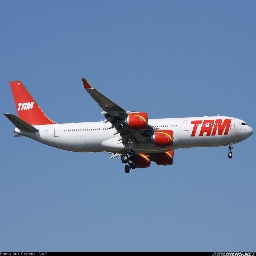}\includegraphics[width=\CrossQualImgSize\textwidth,height=\CrossQualImgSize\textwidth]{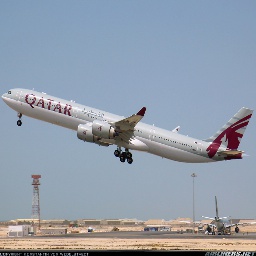}
	\end{tabular}
\caption{Qualitative misclassification interpretation. The odd columns show a misclassified test image while the even columns show the nearest neighbor from the \textit{training} split.}
\label{fig:retrieval_interpretability}
\end{figure}

It is challenging, for the current classification architectures, to interpret a test image misclassification.
By learning image embedding through a secondary head, it becomes trivial to investigate an image's test and train splits neighborhood. Figure~\ref{fig:retrieval_interpretability} shows nine (three images per odd column) misclassified test images and their corresponding nearest neighbor from the train split. The resembles between a misclassified test image and a particular training image can reveal corner cases omitted while collecting the data. One interesting statistic is that 79.34\% of misclassified predictions, from Flowers-102 test split, match the label of their nearest training neighbor.   This emphasizes the classification complexity level of FGVR.


\subsection{Ablation Analysis}
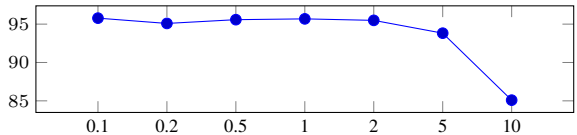
\begin{figure}[t]
	\scriptsize
	\begin{tikzpicture} \begin{axis}[width=0.5\textwidth, height=3.0cm, enlargelimits=0.15, xtick=data, nodes near coords align={vertical},
	symbolic x coords={0.1,0.2,0.5,1,2,5,10}] 
	\addplot coordinates {
		(0.1,95.78) (0.2,95.09) (0.5,95.58) (1,95.685) (2,95.49) (5,93.82) (10,85.09)
	};  
	\end{axis}
	\end{tikzpicture}
	\caption{Hyper-parameter $\lambda$ tuning on the Flowers-102 dataset.}
	\label{fig:hyper_tune}
\end{figure}

\noindent\underline{\textbf{Hyper-Parameter Stability:}} Our approach has two hyper-parameters: $\lambda$ and the embedding dimensionality $d_{\text{emb}}$. $\lambda$ is tuned on the Flowers-102 dataset through the validation split. All hyper-parameter tuning experiments are executed for 2000 iterations. Figure~\ref{fig:hyper_tune} highlights $\lambda$ stability within $[0.1,2]$. A larger $\lambda$ making triplet loss dominant is discouraged. Intuitively, further hyper-parameters tuning can achieves better performance.

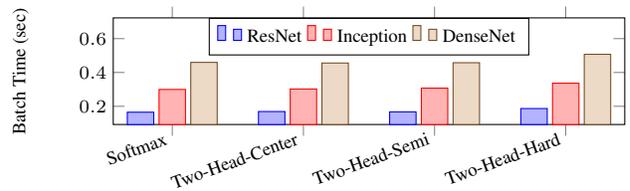
\begin{figure}[t]
	\scriptsize
	\begin{tikzpicture} \begin{axis}[ybar,width=0.48\textwidth, height=3.0cm, enlargelimits=0.15, xtick=data, nodes near coords align={vertical},
	ylabel={Batch Time (sec)},
	ymax=0.65,
	symbolic x coords={Softmax,Two-Head-Center,Two-Head-Semi,Two-Head-Hard},
	x tick label style={rotate=20,anchor=east},
	legend style={at={(0.5,1.0)}, anchor=north,legend columns=-1},
	] 
	\addplot coordinates {
		(Softmax,0.1650968) (Two-Head-Center,0.1690896) (Two-Head-Semi,0.1670304) (Two-Head-Hard,0.1867746)
	};
	\addplot coordinates {
		(Softmax,0.2996722) (Two-Head-Center,0.3023132) (Two-Head-Semi,0.3075474) (Two-Head-Hard,0.3366508)
	};

	\addplot coordinates {
		(Softmax,0.4596562) (Two-Head-Center,0.4559488) (Two-Head-Semi,0.4576844) (Two-Head-Hard,0.5074326)
	};      
	\legend{ResNet,Inception,DenseNet}
	\end{axis}
	\end{tikzpicture}
	\caption{Two-head time complexity analysis on ResNet-50, Inception-V4 and DenseNet-161 using Flowers-102 dataset.}
	\label{fig:time_performance}
\end{figure}

\noindent\textbf{\underline{Two-Head Time Complexity:}} The computational cost of the embedding head is negligible. Both sampling and backpropagation are implemented on GPU. Training time increases by $1\%, 3\%,$ and $2\%$ for semi-hard, hard and center losses on Titan XP GPU, respectively.  Figure~\ref{fig:time_performance} shows a time complexity analysis in terms of batch processing time (secs). Please note that triplet loss approaches retain from computing classes centers or enforcing a specific number of modes.

\subsection{Discussion}
Our experiments demonstrate how a two-head architecture with triplet loss outperforms a vanilla single-head softmax network. Triplet loss attains the center loss, triplet center loss and magnet loss objectives without enforcing explicit class representatives. It promotes both intra-class compactness and inter-class margin maximization. Semi-hard triplet loss relaxes the unimodal embedding constraint while maintaining stabler learning curve. Hard triplet loss achieves larger improvement margins but can suffer model collapse. Triplet loss effectively regularizes softmax and promote better feature embedding.


The two-head architecture with triplet loss is the main scope of this paper. Investigating other recent ranking losses, \eg Margin loss~\cite{wu2017sampling}, and comparing their benefits to softmax remains an open question.


\section{Conclusion}
We propose a seamless integration of  triplet loss as an embedding regularizer into standard classification architectures. The regularizer competence is illustrated on multiple datasets, architectures and recognition tasks. Triplet loss, without the large batch requirement, boosts standard architectures' performance. With minimal hyper-parameter tuning and a \textit{single} fully connected layer on top of pretrained standard architectures, we promote generality to novel domains. Promising results are achieved on an imbalanced dataset. We incur a minimal computational overhead during training, but raise classification model efficiency and interpretability. Our architectural extension enables both retrieval and classification tasks during inference.

{\small
\bibliographystyle{ieee}
\bibliography{egbib}
}
\clearpage

\newcommand{\beginsupplement}{%
	\setcounter{table}{0}
	\renewcommand{\thetable}{S\arabic{table}}%
	\setcounter{figure}{0}
	\renewcommand{\thefigure}{S\arabic{figure}}%
}

\beginsupplement

\section{Supplementary Material}
The next subsections provide more details about our architecture and training procedure's technicalities. Further quantitative evaluations on fine-grained visual recognition (FGVR) are presented. Finally, we demonstrate the training procedure for the Honda Research Institute Driving Dataset.

\subsection{Fine-Grained Visual Recognition}
\begin{figure}[t]
	\centering
	\includegraphics[width=0.7\linewidth]{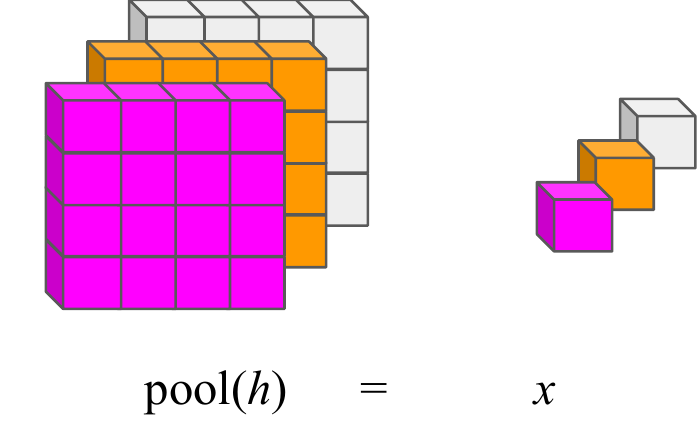}
	\caption{Orderless pooling reduces dimensionality but loses features spatial information.}
	\label{fig:pool_arch}
\end{figure}

Figure 2 in the main paper presents our two-head architecture. The pre-logit layer $x$ supports the softmax loss. Similarly, triplet loss utilizes $h$, where $x = pool(h)$. The network outputs, both logits and embedding, are formulated as follows.
\begin{align}
\text{logits} &= W_{\text{logits}} * \text{flat}(x) \\
\text{embedding} &= W_{\text{emb}} * \text{flat}(h). \label{eq:w_emd}
\end{align}
Orderless pooling, like averaging, reduces $h$ dimensionality but loses spatial information. For example, in DenseNet161, $h \in R^{7\times7\times2208}$ while $x \in R^{1\times1\times2208}$. Thus, $W_{\text{emb}}$ employs $h$, instead of $x$, to improve feature embedding. Figure~\ref{fig:pool_arch} illustrates how $h$ provides a finer control level while learning $W_{\text{emb}}$.

\begin{figure}[t]
	\centering
	\includegraphics[width=0.7\linewidth]{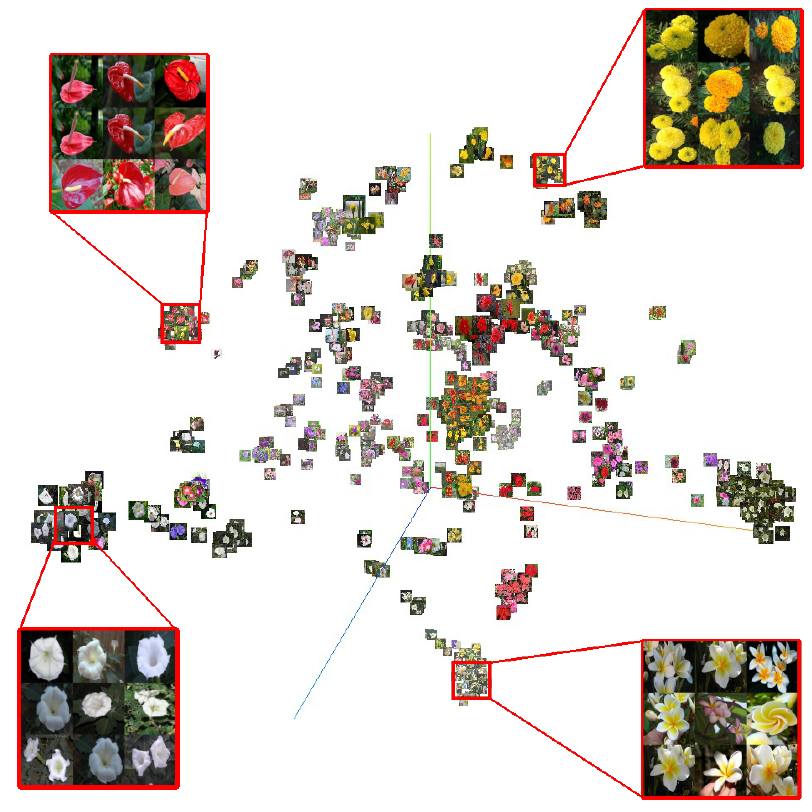}
	\caption{t-SNE visualization for Flowers-102 embedding using 50 random classes, 20 samples per class. Best viewed in color.}
	\label{fig:tsne}
\end{figure}

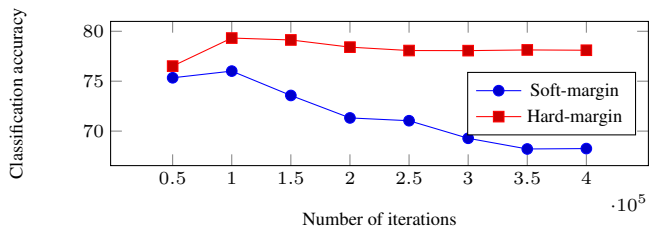
\begin{figure}[t]
	\scriptsize
	\begin{tikzpicture} \begin{axis}[width=0.5\textwidth, height=3.5cm, enlargelimits=0.15, xtick=data, nodes near coords align={vertical},
	legend style={at={(0.82,0.65)}, anchor=north},
	ylabel={Classification accuracy},
	xlabel={Number of iterations},
	] 
	\addplot coordinates {
		(50000,75.34) (100000,76.01) (150000,73.57) (200000,71.32) 
		(250000,71.04) (300000,69.28) (350000,68.21) (400000,68.25) 
	};  
	\addplot coordinates {
		(50000,76.51) (100000,79.32) (150000,79.13) (200000,78.41) 
		(250000,78.07) (300000,78.06) (350000,78.13) (400000,78.10) 
	};  
	\legend{Soft-margin,Hard-margin}
	\end{axis}
	\end{tikzpicture}
	\caption{Model collapse study by training DenseNet161 for 400K iterations. Triplet loss with hard-mining evaluated with soft and hard margins.}
	\label{fig:mode_collapse}
\end{figure}

Figure~\ref{fig:tsne} shows a t-SNE visualization for Flowers-102 embedding  using 50 random classes, 20 samples per class.  In the main paper, the inferior performance of triplet loss with hard-mining is associated with convergence to bad local minima, \ie, a collapsed model $(\ie f(x) = 0)$~\cite{schroff2015facenet}. To examine such assumption, we train a DenseNet for 400K iterations on Stanford Dogs. This large number of iterations increases the chances of a model collapse. Figure~\ref{fig:mode_collapse} presents the performance on the test split after every 50K iterations. Triplet loss with hard-mining is evaluated with both soft and hard margin. Soft margin applies the softplus function $\ln(1+\exp( \sbullet[.75] ))$ while hard margin uses a fixed margin  $m=0.2$. The triplet loss with hard-mining deteriorates with soft margin when trained for a large number of iterations. Hard-mining with hard margin is more robust. We found similar behavior on other datasets like Stanford Cars and Aircrafts datasets.

Table 5 in the main paper presents a quantitative analysis for the feature embedding learned by the second head in our proposed architecture. Similarly, Table~\ref{tbl:penultimate_embedding_qual} presents feature embedding quantitative analysis using the architecture penultimate layer, \ie, layer $x$ (Figure 2 in the main paper). This layer is present in both our proposed two-head and single-head (vanilla softmax) architecture. Similar to Table 5, the triplet loss embedding is superior to the softmax embedding. Triplet loss with hard-mining achieves the best results on ResNet-50 but degrades on Inception-V4 trained for 80K iterations. Center loss achieves good results with DenseNet161 on NABirds but generally fluctuates and suffers with Inception-V4. Triplet loss with semi-hard margin achieves sub-optimal embedding but maintains the highest stability compared to center and hard-mining approaches.

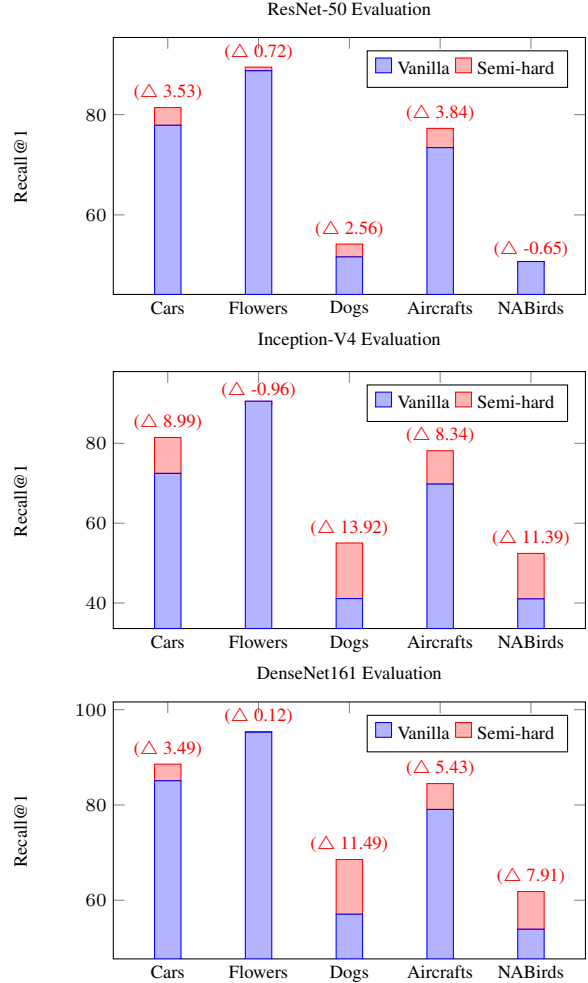
\begin{figure}[!h]
	\scriptsize
	\centering
	\begin{tikzpicture} 
	\begin{axis}[ybar stacked,width=0.45\textwidth,height=5cm, enlargelimits=0.15, 
	title=ResNet-50 Evaluation,
	xtick=data, 
	ylabel={Recall@1},
	symbolic x coords={Cars,Flowers,Dogs,Aircrafts,NABirds},
	legend style={at={(0.75,0.95)}, anchor=north,legend columns=-1},
	nodes near coords,
	nodes near coords align={vertical},
	] 
	\addplot+[point meta=explicit symbolic] coordinates {
		(Cars,77.88) (Flowers,88.76) (Dogs,51.62) (Aircrafts,73.42) (NABirds,50.70)
	};
  \addplot+[point meta=explicit symbolic] coordinates {
	
		(Cars,3.53) [($\triangle$ 3.53)]
	(Flowers,0.72) [($\triangle$ 0.72)]
	(Dogs,2.56)  [($\triangle$ 2.56)]
	(Aircrafts,3.84)  [($\triangle$ 3.84)]
	(NABirds,-0.65) [($\triangle$ -0.65)]
	
  };

	\legend{Vanilla,Semi-hard}
	\end{axis}
	\end{tikzpicture}
	\begin{tikzpicture} \begin{axis}[ybar stacked,width=0.45\textwidth,height=5cm,
	title=Inception-V4 Evaluation,
	enlargelimits=0.15, 
	xtick=data, 
	ylabel={Recall@1},
	symbolic x coords={Cars,Flowers,Dogs,Aircrafts,NABirds},
	legend style={at={(0.75,0.95)}, anchor=north,legend columns=-1},
	nodes near coords,
	nodes near coords align={vertical},
	] 
	\addplot+[point meta=explicit symbolic] coordinates {
		(Cars,72.47) (Flowers,90.54) (Dogs,41.11) (Aircrafts,69.79) (NABirds,41.03)
	};

\pgfmathsetmacro{\cars}{8.99}
\pgfmathsetmacro{\flowers}{-0.96}
\pgfmathsetmacro{\dogs}{13.92}
\pgfmathsetmacro{\aircrafts}{8.34}
\pgfmathsetmacro{\nabirds}{11.39}
	\addplot+[point meta=explicit symbolic] coordinates {
		(Cars,\cars) [($\triangle$ \cars)]
		(Flowers,\flowers) [($\triangle$ \flowers)]
		(Dogs,\dogs)  [($\triangle$ \dogs)]
		(Aircrafts,\aircrafts)  [($\triangle$ \aircrafts)]
		(NABirds,\nabirds) [($\triangle$ \nabirds)]
	};
	
	\legend{Vanilla,Semi-hard}
	\end{axis}
	\end{tikzpicture}

	\begin{tikzpicture} \begin{axis}[ybar stacked,width=0.45\textwidth,height=5cm, enlargelimits=0.15, 
	title=DenseNet161 Evaluation,
	xtick=data, 
	ylabel={Recall@1},
	symbolic x coords={Cars,Flowers,Dogs,Aircrafts,NABirds},
	legend style={at={(0.75,0.95)}, anchor=north,legend columns=-1},
	nodes near coords,
	nodes near coords align={vertical},
	] 
	\addplot+[point meta=explicit symbolic] coordinates {
		(Cars,85.08) (Flowers,95.28) (Dogs,57.06) (Aircrafts,79.06) (NABirds,53.91)
	};
	
	\pgfmathsetmacro{\cars}{3.49}
	\pgfmathsetmacro{\flowers}{0.12}
	\pgfmathsetmacro{\dogs}{11.49}
	\pgfmathsetmacro{\aircrafts}{5.43}
	\pgfmathsetmacro{\nabirds}{7.91}
	
	\addplot+[point meta=explicit symbolic] coordinates {
		(Cars,\cars) [($\triangle$ \cars)]
		(Flowers,\flowers) [($\triangle$ \flowers)]
		(Dogs,\dogs)  [($\triangle$ \dogs)]
		(Aircrafts,\aircrafts)  [($\triangle$ \aircrafts)]
		(NABirds,\nabirds) [($\triangle$ \nabirds)]
	};
	
	\legend{Vanilla,Semi-hard}
	\end{axis}
	\end{tikzpicture}
\caption{Comparative embedding evaluation between single-head softmax and two-head with semi-hard triplet loss using the penultimate layer in ResNet-50, Inception-V4 and DenseNet161 respectively. Triplet loss semi-hard improvements over the softmax model are reported as ($\triangle$).}
	\label{fig:com_recall_eval}
\end{figure}

Figure~\ref{fig:com_recall_eval} graphically summarizes Table~\ref{tbl:penultimate_embedding_qual}. It provides a comparative embedding evaluation between the single-head softmax verses the two-head with semi-hard triplet loss using recall@1 metric. Triplet loss improvements, over the softmax model, are reported as ($\triangle$). The Flowers-102 dataset has the smallest training split with 1020 images only. With this limited data, the head-two architecture achieves marginal improvement if any.

Table~\ref{tbl:fgvr_densenet} compares our proposed two-head architecture, using DenseNet161, with state-of-the-art approaches on the five FGVR datasets. Our two-head architecture with the semi-hard triplet loss regularizer achieves competitive results.

\begin{table}[!h]
	\scriptsize
	\centering
	\begin{tabular}{@{}llccccc@{}}
		&      & NMI   & R@1     & R@4   & R@8   & R@16  \\ \hline
		
		\multirow{4}{*}{Cars - ResNet} 
		& Vanilla & 0.791 & 77.88 & 91.17 & 94.65 & 96.9  \\
		& CNTR    & 0.756 & 77.98 & 91.12 & 94.58 & 96.78 \\
		& SEMI    & 0.823 & 81.41 & 92.79 & 95.91 & 97.74 \\
		& HARD    & \textbf{0.853} & \textbf{85.31} & \textbf{94.30}  & \textbf{96.82} & \textbf{98.07} \\ \hline
		
		\multirow{4}{*}{Flowers - ResNet} 
		& Vanilla & 0.800 & 88.76 & 95.51 & 97.27 & \textbf{98.49} \\
		& CNTR    & 0.807 & 88.79 & 95.58 & 97.32 & \textbf{98.49} \\
		& SEMI    & \textbf{0.818} & 89.48 & \textbf{95.82} & \textbf{97.37} & 98.37 \\
		& HARD    & 0.742 & \textbf{90.78} & 95.56 & 96.93 & 97.98 \\ \hline
		
		\multirow{4}{*}{Dogs - ResNet} 
		& Vanilla & 0.587 & 51.62 & 74.22 & 83.02 & 89.76 \\
		& CNTR    & 0.526 & 48.74 & 71.90 & 80.92 & 87.81 \\
		& SEMI    & 0.621 & 54.18 & 76.39 & 84.50 & 91.10 \\
		& HARD    & \textbf{0.684} & \textbf{60.37} & \textbf{80.34} & \textbf{87.33} & \textbf{92.26} \\ \hline

		\multirow{4}{*}{Aircrafts - ResNet} 
		& Vanilla & 0.756 & 73.42 & 87.88 & 92.26 & 94.90 \\
		& CNTR    & 0.677 & 70.84 & 85.84 & 90.79 & 93.91 \\
		& SEMI    & 0.792 & 77.26 & 89.65 & 93.07 & 95.29 \\
		& HARD    & \textbf{0.829} & \textbf{84.01} & \textbf{91.63} & \textbf{94.21} & \textbf{95.65} \\ \hline
		
		\multirow{4}{*}{NABirds - ResNet} & Vanilla & 0.669 & 50.70 & 71.20 & 79.48 & 85.80 \\
		& CNTR    & 0.623 & 47.40 & 68.18 & 76.56 & 83.33 \\
		& SEMI    & 0.657 & 50.05 & 70.83 & 78.84 & 85.52 \\
		& HARD    & \textbf{0.723} & \textbf{55.85} & \textbf{75.81} & \textbf{83.26} & \textbf{88.67} \\ \hline \hline
		
		\multirow{4}{*}{Cars - Inc-V4} & Vanilla & 0.660 & 72.47 & 86.77 & 90.55 & 93.55 \\
		& CNTR    & 0.496 & 61.55 & 79.09 & 85.09 & 89.69 \\
		& SEMI    & \textbf{0.788} & \textbf{81.46} & \textbf{92.14} & \textbf{94.64} & \textbf{96.37} \\
		& HARD    & 0.566 & 63.70  & 82.04 & 87.54 & 91.42 \\ \hline
		
		\multirow{4}{*}{Flowers - Inc-V4} & Vanilla & 0.778 & 90.54 & 96.21 & 97.63 & 98.70 \\
		& CNTR    & 0.707 & 85.62 & 93.74 & 95.95 & 97.56 \\
		& SEMI    & \textbf{0.801} & 89.58 & 95.23 & 96.91 & 97.84 \\
		& HARD    & 0.731 & \textbf{92.68} & \textbf{96.21} & \textbf{97.27} & \textbf{98.32} \\ \hline
		
		\multirow{4}{*}{Dogs - Inc-V4} 
		& Vanilla & 0.421 & 41.11 & 62.97 & 72.59 & 81.13 \\
		& CNTR    & 0.453 & \textbf{57.13} & 68.32 & 72.35 & 76.90 \\
		& SEMI    & \textbf{0.609} & 55.03 & \textbf{76.50} & \textbf{84.44} & \textbf{90.23} \\
		& HARD    & 0.330 & 33.89 & 54.28 & 65.06 & 74.98 \\ \hline

		\multirow{4}{*}{Aircrafts - Inc-V4} 
		& Vanilla & 0.680 & 69.79 & 85.18 & 89.23 & 91.93 \\
		& CNTR    & 0.546 & 61.60 & 79.75 & 85.33 & 89.53 \\
		& SEMI    & 0.751 & 78.13 & 89.20 & 91.78 & 94.27 \\
		& HARD    & \textbf{0.831} & \textbf{86.26} & \textbf{91.87} & \textbf{93.49} & \textbf{94.72} \\ \hline
		
		\multirow{4}{*}{NABirds - Inc-V4} & Vanilla & 0.546 & 41.03 & 60.11 & 68.88 & 76.71 \\
		& CNTR    & 0.438 & 24.30 & 40.43 & 49.38 & 58.78 \\
		& SEMI    & \textbf{0.638} & \textbf{52.42} & \textbf{72.38} & \textbf{79.57} & \textbf{85.60}  \\
		& HARD    & 0.433 & 23.68 & 38.95 & 47.48 & 57.10 \\ \hline \hline
		
		\multirow{4}{*}{Cars - Dense} & Vanilla & 0.813 & 85.08 & 94.49 & 96.84 & 98.22 \\
		& CNTR    & 0.787 & 87.39 & 93.17 & 94.64 & 95.97 \\
		& SEMI    & 0.875 & 88.57 & 96.08 & 97.66 & 98.71 \\
		& HARD    & \textbf{0.892} & \textbf{89.44} & \textbf{96.38} & \textbf{97.86} & \textbf{98.76} \\ \hline
		
		\multirow{4}{*}{Flowers - Dense} & Vanilla & 0.838 & 95.28 & 98.23 & 98.94 & 99.38 \\
		& CNTR    & 0.812 & 95.87 & 98.16 & 98.75 & 99.22 \\
		& SEMI    & 0.864 & 95.40 & 98.39 & 99.09 & 99.46 \\
		& HARD    & \textbf{0.865} & \textbf{95.79} & \textbf{98.50} & \textbf{99.14} & \textbf{99.50}  \\ \hline
		
		\multirow{4}{*}{Dogs - Dense} & Vanilla & 0.544 & 57.06 & 78.72 & 85.98 & 91.84 \\
		& CNTR    & 0.720 & \textbf{70.96} & 84.00 & 88.19 & 91.96 \\
		& SEMI    & 0.728 & 68.55 & 87.04 & 92.18 & 95.83 \\
		& HARD    & \textbf{0.756} & 70.63 & \textbf{87.80} & \textbf{92.95} & \textbf{96.22} \\ \hline
		
		\multirow{4}{*}{Aircrafts - Dense} & Vanilla & 0.768 & 79.06 & 91.66 & 94.66 & 96.49 \\
		& CNTR    & 0.792 & 86.20 & 91.63 & 93.16 & 94.48 \\
		& SEMI    & 0.853 & 84.49 & \textbf{94.15} & 95.68 & \textbf{96.97} \\
		& HARD    & \textbf{0.856} & \textbf{85.51} & 93.70 & \textbf{95.83} & 96.94	\\ \hline
		
		\multirow{4}{*}{NABirds - Dense} 
		& Vanilla & 0.606 & 53.91 & 73.08 & 80.70 & 86.44 \\
		& CNTR    & \textbf{0.818} & \textbf{75.28} & \textbf{86.88} & \textbf{90.85} & \textbf{93.69} \\
		& SEMI    & 0.677 & 61.82 & 80.70 & 87.07 & 91.62 \\
		& HARD    & 0.674 & 61.64 & 80.21 & 86.77 & 91.37 \\ \hline \hline
	\end{tabular}
	\caption{Detailed feature embedding quantitative analysis across the five datasets using ResNet-50, Inception-V4 and DenseNet-161 architectures' penultimate layer $x$. Triplet with hard mining achieves superior embedding with ResNet-50 trained for 40K iterations. Semi-hard triplet is competitive and stable with Inception-V4 trained for 80K iterations. Center loss suffers a high instability.}
	\label{tbl:penultimate_embedding_qual}
\end{table}

\begin{table*}[t]
	\centering
	\scriptsize
	\parbox{.18\linewidth}{
		\centering
		Flowers-102\\
		\begin{tabular}{@{}lc@{}}
			\hline
			Method          & Acc \\ \hline
			Det.+Seg.~\cite{angelova2013efficient}    &   80.66    \\ 
			Overfeat~\cite{sharif2014cnn}      &   86.80   \\  \hline
			Softmax         &   92.56   \\ 
			\textbf{Two-Head (Semi)} &  \textbf{93.65}   \\ \hline
		\end{tabular}	
	}
	\hfill
	\parbox{.18\linewidth}{
		\centering
		Aircrafts\\
		\begin{tabular}{@{}lc@{}}
			\hline
			Method          & Acc \\ \hline
			LRBP ~\cite{kong2017low}    &   87.30    \\ 
			Liu \etal~\cite{lin2017improved}     &   88.50    \\ \hline
			Softmax         &   89.13   \\ 
			\textbf{Two-Head (Semi)} &  \textbf{89.64}   \\ \hline
		\end{tabular}	
	}
	\hfill
	\parbox{.18\linewidth}{
		\centering
		NABirds\\
		\begin{tabular}{@{}lc@{}}
			\hline
			Method          & Acc \\ \hline
			Branson \etal~\cite{zhang2016weakly}    &   35.70    \\ 
			Van \etal~\cite{krause2016unreasonable}      &   75.00   \\ \hline 
			Softmax         &    78.69  \\ 
			\textbf{Two-Head (Semi)} &   \textbf{79.57}  \\ \hline
		\end{tabular}	
	}
	\hfill
	\parbox{.18\linewidth}{
		\centering
		Cars\\
		\begin{tabular}{@{}lc@{}}
			\hline
			Method          & Acc \\ \hline
			Liu \etal~\cite{liu2016hierarchical}      &   86.80  \\
			Liu \etal~\cite{lin2017improved}      &   92.00   \\ \hline
			Softmax         &    91.64  \\ 
			\textbf{Two-Head (Semi)} &   \textbf{92.36}  \\ \hline
		\end{tabular}	
	}
	\hfill
	\parbox{.18\linewidth}{
		\centering
		Dogs\\
		\begin{tabular}{@{}lc@{}}
			\hline
			Method          & Acc \\ \hline
			Zhang \etal~\cite{zhang2016weakly}      &   80.43  \\
			Krause \etal~\cite{krause2016unreasonable}      &   80.60   \\ \hline
			\textbf{Softmax}         &    \textbf{81.58}  \\ 
			Two-Head (Semi) &   80.89  \\ \hline
		\end{tabular}	
	}
	\caption{Quantitative evaluation on the five FGVR datasets using DenseNet161. Our two-head architecture with semi-hard triplet loss regularizer compares favorably with state-of-the-art results.}
	\label{tbl:fgvr_densenet}
\end{table*}

\subsection{Autonomous Car Driving}

The Honda Research Institute Driving Dataset (HDD) contains 137 sessions $S$. Each session $S_i$ represents a navigation task performed by a driver. $S$ is divided into 93, 5, and 36 sessions for training, validation and testing splits respectively. Three sessions are removed for missing annotations. HDD has four annotation layers to study the drivers' actions: (1) Goal-oriented, (2) stimulus-driven, (3) cause and (4) attention. The \textbf{Goal-oriented} layer, utilized in our experiments , defines drivers' actions to reach their destinations, \eg, \textit{left-turn} and \textit{intersection passing}. Ramanishka \etal~\cite{RamanishkaCVPR2018} provides further details for the other three annotation layers.

Triplet loss mini-batches require both positive and negative samples. The FGVR datasets have uniform class distribution. Thus, training batches' construction is straight-forward by sampling random classes and their corresponding images as outlined in the main paper. On the other hand, HDD suffers class imbalance. A different batch construction procedure is required. 

Algorithm~\ref{alg:sampling} outlines our training procedure. First, $N_B$ mini-batches are constructed, each containing $b$ random actions. The batches' embeddings are computed using $N_B$ feed forward passes. The $2D$ matrix $D_\phi$ stores the pairwise distance between the $N_B\times b$ actions. All positive pairs $(a,p)$ and their corresponding semi-hard negatives $n$ are identified. For a fair comparison with vanilla softmax approach, only $(b//3)$ random triplets $(a,p,n)$ are utilized for back-propagation. This process repeats for $N$ training iterations.


\begin{algorithm}[h]
	\caption{HDD training procedure. In our experiments, $b=\{33,36\}$ is the mini-batch size, $N_B=3$ is the number of mini-batches, and $N=10K$ is number of training iterations.}
	\begin{algorithmic}
		\FORALL{iteration $i$ in N}
			\STATE  $S_\phi = \Phi $ 
			\FORALL{$j$ in $N_B$} 
				\STATE  Add a random batch, of size $b$, to $S_\phi$
			\ENDFOR 
			\STATE Compute action embeddings $E_\phi$ for $S_\phi$
			\STATE Compute pairwise distance matrix $D_\phi$ using $E_\phi$
			\STATE $T_{tri} = \Phi$
			\STATE Construct all positive pairs $(a,p)$
			\FORALL{$(a,p)$ in positive pairs} 
				\STATE Find nearest semi-hard negative $n$	using $D_\phi$ 
				\STATE append $(a,p,n)$ to $T_{tri}$
			\ENDFOR
			\IF {$len(T_{tri}) > b//3 $}
				\STATE $T_{tri}=$shuffle$(T_{tri})[0:b//3]$
			\ENDIF
			\STATE // $T_{tri}$ contains $b$ actions
			\STATE Feed-forward  $T_{tri}$
			\STATE compute softmax + triplet losses and back-propagate.
		\ENDFOR 
	\end{algorithmic}
	\label{alg:sampling}
\end{algorithm}

\end{document}